\documentclass[final,5p,times,twocolumn]{elsarticle}

\graphicspath{{figs1/}}

\usepackage{amssymb}
\usepackage{amsmath}

\usepackage{graphicx}
\usepackage{subcaption}
\usepackage[labelfont=bf]{caption}
\usepackage{float}
\usepackage{hyperref}
\hypersetup{
    colorlinks=true,
    linkcolor=blue,
    filecolor=magenta,      
    urlcolor=blue,
}
\usepackage{placeins}
\usepackage{bm,csquotes}
\usepackage{enumerate}
\usepackage{booktabs}
\usepackage{algorithm}%
\usepackage{algorithmicx}%
\usepackage{algpseudocode}%

\usepackage{siunitx}
\usepackage{multirow}
\usepackage{textcomp}
\usepackage{newunicodechar}
\newunicodechar{′}{\ensuremath{^{\prime}}}

\makeatletter
\renewcommand\paragraph{\@startsection{paragraph}{4}{\z@}%
  {1ex \@plus 1ex \@minus .2ex}%
  {-1em}%
  {\normalfont\normalsize\bfseries}}
\makeatother

\journal{Artificial Intelligence In Medicine}

\begin{document}

\begin{frontmatter}



\title{Unsupervised Consensus‑Based Anomaly Detection for Spatiotemporal Malaria Incidence in Ghana}

\author[label2]{T. Ansah-Narh\corref{cor1}}
\cortext[cor1]{Corresponding author.}
\ead{theophilus.ansah-narh@gaec.gov.gh}

\author[label1]{Y. Asare Afrane} 




\affiliation[label2]{organization={Ghana Space Science and Technology Institute, Ghana Atomic Energy Commission},
            addressline={P. O. Box LG 80}, 
            city={Legon},
            state={Accra},
            country={Ghana}}
            
\affiliation[label1]{organization={Department of Medical Microbiology, University of Ghana Medical School}, 
            city={University of Ghana},
            state={Accra},
            country={Ghana}}




\begin{abstract}
Routine malaria surveillance systems are designed to monitor disease burden and seasonal trends, yet they provide limited insight into when and where transmission departs from expected behaviour. Identifying such anomalies is important because unusual transmission episodes may signal emerging risks that are not apparent from case counts alone. This study developed a consensus anomaly detection framework and applied it to monthly malaria surveillance records from Ghana between 2014 and 2023 to identify atypical transmission patterns across space and time.
The analysis revealed that malaria anomalies were highly structured geographically and temporally rather than randomly distributed throughout the surveillance record. Ashanti and Northern Regions accounted for the majority of recurrent anomalies, while district-level analyses identified persistent hotspot behaviour centred on Tamale, Kumasi, and the Accra metropolitan area. A notable finding was that anomaly burden and anomaly frequency were spatially distinct. Tamale contributed the largest malaria burden during anomalous periods, whereas the highest anomaly rates were concentrated within a cluster of districts in Ashanti Region, indicating that areas with the greatest malaria burden were not necessarily those experiencing the most persistent anomalous behaviour. Anomalous months also formed a statistically distinct epidemiological population, exhibiting markedly higher malaria burden than normal months (Cohen's $d=3.252$) together with large departures from seasonal expectations and regional transmission baselines ($d>1.2$).
These findings demonstrate that malaria burden alone provides an incomplete picture of transmission dynamics. By distinguishing locations where malaria is most prevalent from those where transmission behaves most unusually, the study reveals previously hidden spatial and temporal dimensions of malaria risk. This anomaly-based perspective provides a complementary framework for strengthening surveillance, prioritising investigations, and supporting more targeted malaria control strategies.
\end{abstract}


\begin{highlights}
\item Consensus anomaly detection identified atypical malaria transmission patterns.
\item Ashanti and Northern Regions accounted for most recurrent anomalies.
\item Tamale showed the largest anomaly burden during transmission extremes.
\item Highest anomaly rates occurred within a cluster of Ashanti districts.
\item Advocates for advanced tools to manage complex insurance data effectively.
\item Malaria burden and anomaly persistence emerged as distinct risk dimensions.
\end{highlights}


\end{frontmatter}



\section{Introduction}
\label{sec1}

Malaria continues to place a substantial burden on public health systems across sub-Saharan Africa. In Ghana, the disease remains one of the leading causes of outpatient attendances and hospital admissions, with approximately 10.2 million suspected cases recorded in outpatient departments during 2018 alone \cite{Agbemafle2023}. Despite sustained control efforts, including insecticide-treated net distribution, indoor residual spraying, and seasonal malaria chemoprevention, malaria transmission continues to exhibit marked spatiotemporal heterogeneity \cite{Agbemafle2023}. The prevalence of malaria infection among children aged 6--59 months has declined from $40.6\,\%$ in 2014 to $8.6\,\%$ in 2022, yet rural areas continue to report prevalence rates approximately three times higher than urban settings. Seasonal transmission drives much of this variability, with peaks typically coinciding with the rainy season, although the strength and timing of seasonal patterns differ markedly between rural, peri-urban and urban districts \cite{Savi2024}. Understanding where and when malaria transmission deviates from expected behaviour is therefore essential for strengthening malaria surveillance and supporting evidence-based public health decision-making. Earlier recognition of unusual transmission patterns may assist health authorities in prioritising epidemiological investigations, assessing surveillance performance, and informing the allocation of diagnostic and treatment resources where operational capacity permits.

Routine malaria surveillance in Ghana relies largely on summary statistics and threshold-based alert systems. The District Health Management Information System (DHIMS2) aggregates facility-level data, and the National Malaria Control Programme uses these data to monitor disease trends, distribute commodities and evaluate malaria control interventions. However, conventional surveillance approaches have important limitations. An evaluation of the surveillance system in the Adaklu District demonstrated that, although the system was simple, acceptable and representative, it failed to detect an epidemic in August 2016 because no alert or epidemic thresholds had been established \cite{Agbemafle2023}. More generally, thresholds based solely on historical means or standard deviations cannot readily adapt to evolving transmission dynamics and are insensitive to multivariate epidemiological patterns, such as moderate increases in malaria incidence accompanied by unusual age distributions or sustained transmission outside the expected seasonal period. Aggregate annual or semi-annual summaries may also obscure relatively short-lived transmission events that warrant further epidemiological investigation. These limitations are not unique to Ghana; similar challenges have been documented across malaria-endemic settings where routine surveillance data remain underutilised for supporting timely epidemiological assessment and decision-making \cite{Eze2023}. Consequently, there is growing interest in analytical approaches that extract earlier and more informative signals from routine surveillance data. Forecasting-based malaria early warning systems have demonstrated that routinely collected surveillance data can support prospective identification of areas at elevated transmission risk and provide evidence for operational planning and disease control interventions \citep{Colborn2018}. However, these approaches are primarily designed to forecast expected transmission patterns rather than identify statistically unusual observations. Anomaly detection addresses this complementary surveillance objective by identifying departures from expected epidemiological behaviour that warrant further investigation.

Unsupervised machine learning provides one such complementary analytical approach because it does not require predefined thresholds or labelled outbreak data. Anomaly detection algorithms identify observations that differ substantially from the dominant behaviour of the data without requiring prior knowledge of what constitutes an outbreak. Within infectious disease surveillance, however, statistical anomalies should not be interpreted automatically as disease outbreaks. Rather, they represent unusual observations that may arise from emerging transmission events, unexpected changes in reporting behaviour, intervention effects, environmental variability, or other departures from expected epidemiological conditions, thereby warranting further investigation. A growing body of work has demonstrated the utility of anomaly detection for infectious disease surveillance. In the Brazilian Amazon, unsupervised anomaly detection models have provided early indications of outbreak onset, transmission peaks and changes in the proportion of positive malaria cases \cite{Eze2023}. In Thailand, clustering-based and time-series anomaly detection methods have been developed to identify unusual malaria activity at the provincial level and integrated into national surveillance dashboards \cite{Srimokla2024}. Similar approaches have also been applied to respiratory disease surveillance using primary healthcare records and transformer-based frameworks for COVID-19 monitoring \cite{hashemi2024surveillance}. Among the most widely used algorithms are Isolation Forest, which isolates anomalous observations by exploiting the property that anomalies are relatively few and distinct; Local Outlier Factor, which identifies observations with substantially lower local density than their neighbours; autoencoders, which detect anomalies through elevated reconstruction error after learning a compressed representation of normal behaviour; and Elliptic Envelope, which identifies observations with low probability under an assumed multivariate Gaussian distribution. Because malaria transmission exhibits multiple temporal behaviours, including abrupt surges, persistent departures from seasonal expectations, gradual changes in long-term transmission intensity, and age-specific variations in disease burden, no single anomaly detection algorithm can be expected to capture every epidemiologically relevant pattern consistently. Consensus approaches that integrate multiple algorithms therefore reduce model-specific bias, improve robustness, and provide a more reliable characterisation of statistically unusual transmission behaviour.

Recently, \citet{AnsahNarh2026} developed a Bayesian nonlinear inference framework to model malaria dynamics in Ghana by integrating a cubic deterministic baseline with a damped oscillatory kernel estimated using an ensemble Markov Chain Monte Carlo sampler. That study demonstrated strong empirical adequacy for both under-five and older age groups and provided probabilistic forecasts of malaria admissions from 2024 to 2026. However, the Bayesian framework assumes a parametric representation of the underlying transmission dynamics and is designed primarily for forecasting aggregate trajectories. It does not automatically identify which specific region-month combinations exhibit unusual behaviour relative to their historical transmission patterns, nor does it provide a systematic mechanism for ranking locations according to the strength of their deviation. Unsupervised anomaly detection addresses a complementary surveillance objective by providing a data-driven, non-parametric framework for identifying observations that depart substantially from expected behaviour without imposing a predefined dynamic structure. By highlighting anomalous region-month combinations, anomaly detection generates statistical signals that may indicate unexpected transmission events which forecasting models may otherwise treat as residual variation. The two approaches are therefore complementary: Bayesian inference characterises expected transmission trajectories together with associated uncertainty, whereas unsupervised anomaly detection identifies statistically significant departures from those expectations.

The present study therefore develops and validates a consensus-based unsupervised anomaly detection framework for regional malaria surveillance data from Ghana covering the period 2014--2023. The central research question is as follows: which region-month combinations exhibit statistically unusual malaria transmission patterns when evaluated using multiple unsupervised learning algorithms, and how do these anomalies relate to established regional and seasonal transmission characteristics? In this study, anomalies are interpreted as statistically significant departures from the historical behaviour of individual regions rather than definitive evidence of malaria outbreaks. Such departures may arise from abrupt increases in malaria incidence, persistent transmission outside the expected seasonal period, unusual age-specific admission patterns, or other unexpected deviations that warrant further epidemiological investigation.

The motivation for this work stems from the increasing availability of routine malaria surveillance data and the need for analytical tools capable of transforming these data into timely epidemiological intelligence. While forecasting models estimate expected future transmission trajectories, anomaly detection addresses a complementary problem by identifying observations that depart substantially from expected behaviour. The proposed framework is therefore intended as a decision-support tool that strengthens routine malaria surveillance by prioritising statistically unusual observations for further epidemiological assessment rather than replacing existing surveillance systems or public health judgement.

The principal contributions of this study are fourfold. First, it presents the first consensus-based unsupervised anomaly detection framework developed for spatiotemporal malaria surveillance in Ghana. Second, it introduces a comprehensive set of engineered epidemiological features, including seasonal residuals, lag effects, region-specific standardised scores and age-specific admission counts, to characterise multiple dimensions of abnormal malaria transmission. Third, it demonstrates how consensus learning improves the robustness and interpretability of anomaly detection relative to individual unsupervised algorithms. Finally, it provides spatial and temporal anomaly maps that complement conventional malaria surveillance by identifying statistically unusual transmission patterns that may assist public health authorities in prioritising surveillance review, epidemiological investigation and resource planning where operational capacity permits.

The remainder of this paper is organised as follows. Section~\ref{sec:studyarea} describes the study area, the malaria surveillance data obtained from the DHIMS2, the preprocessing procedures adopted for the analysis, and the methodological framework, including feature engineering, anomaly detection algorithms, the consensus anomaly detection strategy, and the statistical validation procedures. Section~\ref{sec:RNA} presents the results, examining the spatiotemporal distribution of malaria anomalies, the characterisation of high-confidence anomaly events, district-level hotspot patterns, and the statistical separation between anomalous and routine transmission conditions. Section~\ref{sec:disc} discusses the epidemiological and methodological implications of the findings, their relevance for malaria surveillance, and the limitations and future directions of the proposed framework. Finally, Section~\ref{sec:conc} summarises the principal findings of the study and highlights their significance for anomaly-based malaria surveillance and disease monitoring.

\section{Materials and Methods}\label{sec:studyarea}

\subsection{Study Area and Data Description} \label{sec:studyarea_description}

Ghana is situated on the Gulf of Guinea in West Africa, between latitudes $4.5^\circ\mathrm{N}$ and $11.5^\circ\mathrm{N}$ and longitudes $3.5^\circ\mathrm{W}$ and $1.5^\circ\mathrm{E}$. The country covers approximately $239{,}460~\mathrm{km}^2$, including $8{,}520~\mathrm{km}^2$ of inland water bodies. Administratively, Ghana is divided into 16 regions and 261 districts. The present study focuses on malaria case surveillance across all regions, with data aggregated from the district level to the regional level to enable a coherent spatiotemporal analysis of anomaly patterns.

\begin{figure*}
\begin{minipage}[H]{\linewidth}
\centering
\includegraphics[width=\textwidth]{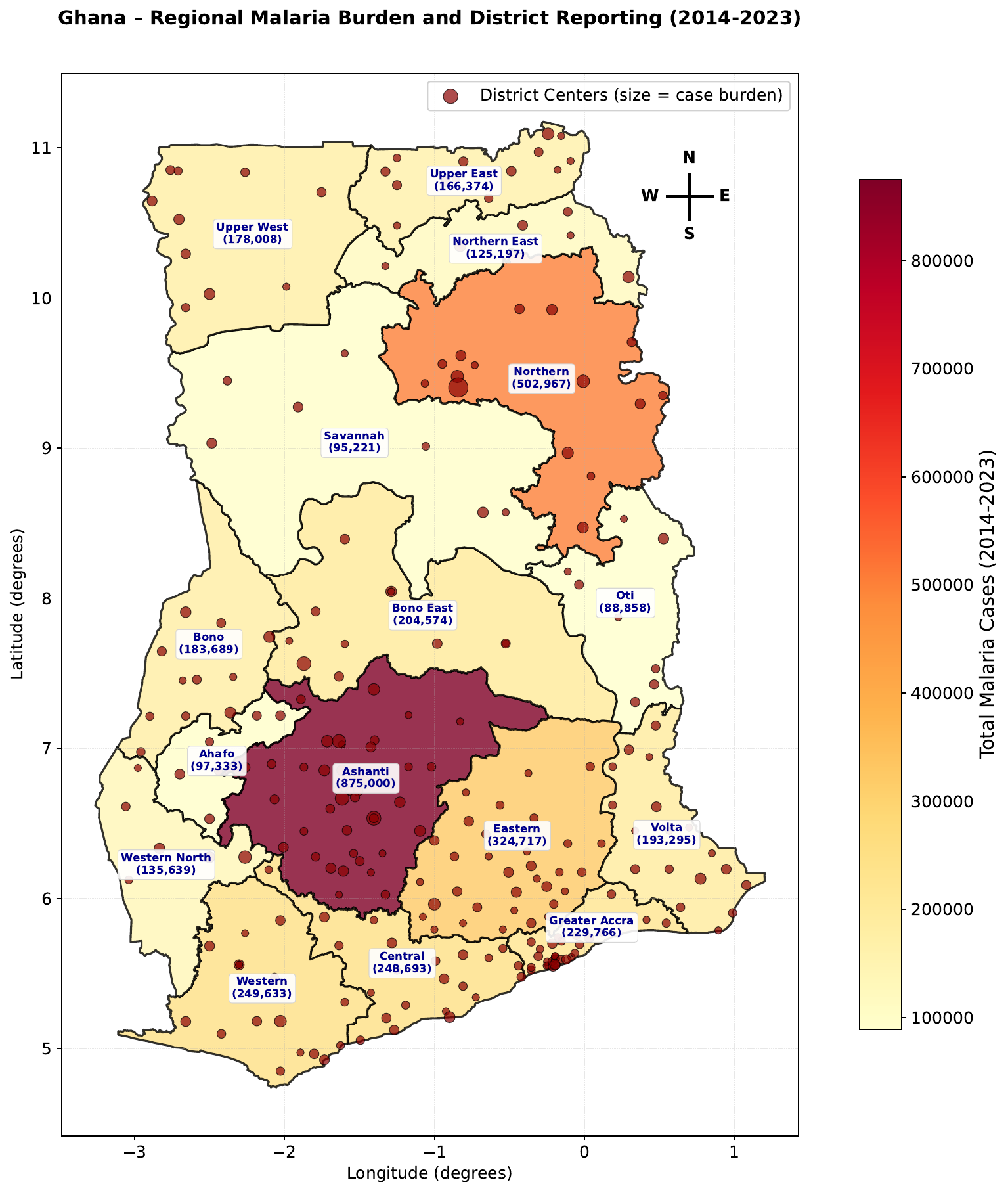} 
\end{minipage}
\caption{Spatial distribution of cumulative malaria admissions across the administrative regions of Ghana during 2014--2023. Regional colour intensity represents the cumulative malaria burden aggregated from district-level records, with darker shades indicating higher numbers of reported admissions. Circles denote district reporting locations, and marker size is proportional to the corresponding cumulative malaria burden. The map highlights substantial spatial heterogeneity in malaria admissions across Ghana, with particularly high burdens observed in the Ashanti and Northern regions.
}\label{fig:ghmap_f1}
\end{figure*}

Figure~\ref{fig:ghmap_f1} presents the study area and the spatial distribution of cumulative malaria burden across Ghana from 2014 to 2023. Regional shading represents the aggregated malaria admissions computed as the sum of reported cases from the two age categories recorded in the surveillance system, namely patients aged less than five years and patients aged five years and above, aggregated across all districts within each region over the study period. District centroids indicate the geographic locations of reporting districts, with marker sizes proportional to the corresponding cumulative malaria burden.
Substantial spatial heterogeneity is evident across the country. The Ashanti Region recorded the highest cumulative malaria burden (875,000 cases), followed by the Northern Region (502,967 cases). Intermediate burdens were observed in the Eastern (324,717 cases), Western (249,633 cases), Central (248,693 cases), and Greater Accra (229,766 cases) regions. In contrast, the lowest cumulative burdens were recorded in Oti (88,858 cases), Savannah (95,221 cases), Ahafo (97,333 cases), Western North (135,639 cases), and North East (125,197 cases).
Overall, the cumulative regional burden varied by nearly an order of magnitude, ranging from approximately 88,900 to 875,000 cases, highlighting pronounced geographical disparities in malaria morbidity across Ghana. These spatial variations provide important context for the subsequent anomaly detection framework and spatiotemporal analysis.

Ghana’s ecological and climatic diversity contributes significantly to spatial heterogeneity in malaria transmission. The country is typically divided into three broad ecological belts: the Guinea Savannah in the north, the Forest zone in the centre, and the Coastal Savannah in the south \cite{adu2015spatiotemporal, deSouza2010}. These zones experience distinct rainfall patterns and hydrological profiles, which shape the seasonality and intensity of malaria transmission. The northern regions generally experience a unimodal rainy season from May to October, while the middle and southern belts have bimodal rainfall peaks (April–June and September–November), influencing vector breeding and transmission cycles accordingly \cite{awine2017towards}.

Malaria remains endemic throughout Ghana, with a consistently high incidence rate reported among children under five years of age, the most vulnerable demographic group \cite{adokiya2025perspectives, aidoo2024malaria, kolekang2022challenges}. Despite ongoing control efforts, the disease exhibits pronounced spatial and seasonal variations, reflecting disparities in vector ecology, healthcare accessibility, and socioeconomic factors across regions. Urban areas, particularly Greater Accra, may report lower prevalence due to improved housing, drainage and health infrastructure, whereas higher transmission persists in rural districts with limited access to preventive and curative services \cite{fobil2011neighborhood}.

\begin{figure*}
\begin{minipage}[H]{\linewidth}
\centering
\includegraphics[width=\textwidth]{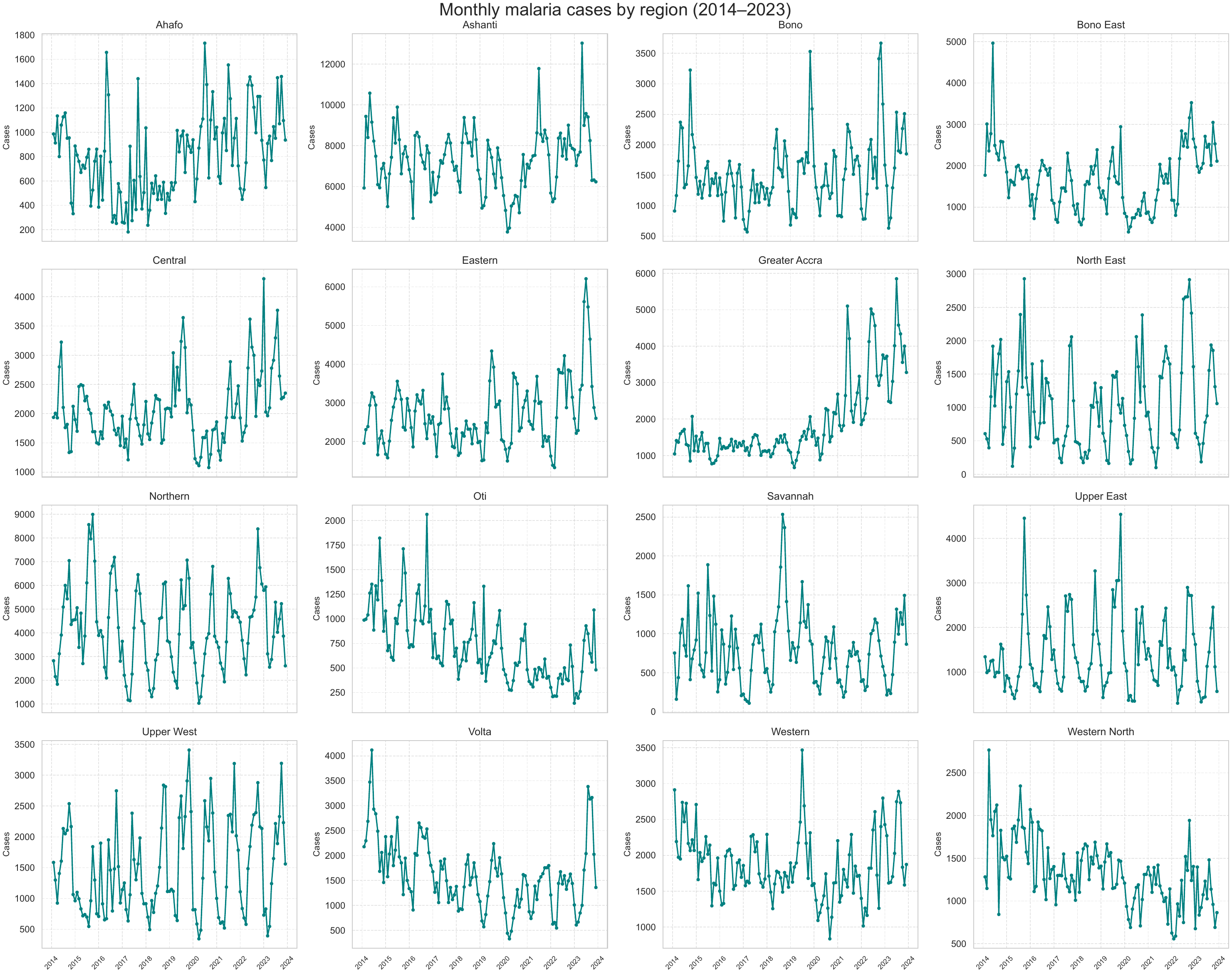} 
\end{minipage}
\caption{Monthly malaria admissions across the 16 administrative regions of Ghana from January 2014 to December 2023. Each panel shows the regional aggregate time series, highlighting substantial spatial heterogeneity in malaria burden, seasonal variability, and long-term temporal dynamics across regions.
}\label{fig:regional_time_series}
\end{figure*}

\begin{figure*}
\begin{minipage}[H]{\linewidth}
\centering
\includegraphics[width=\textwidth]{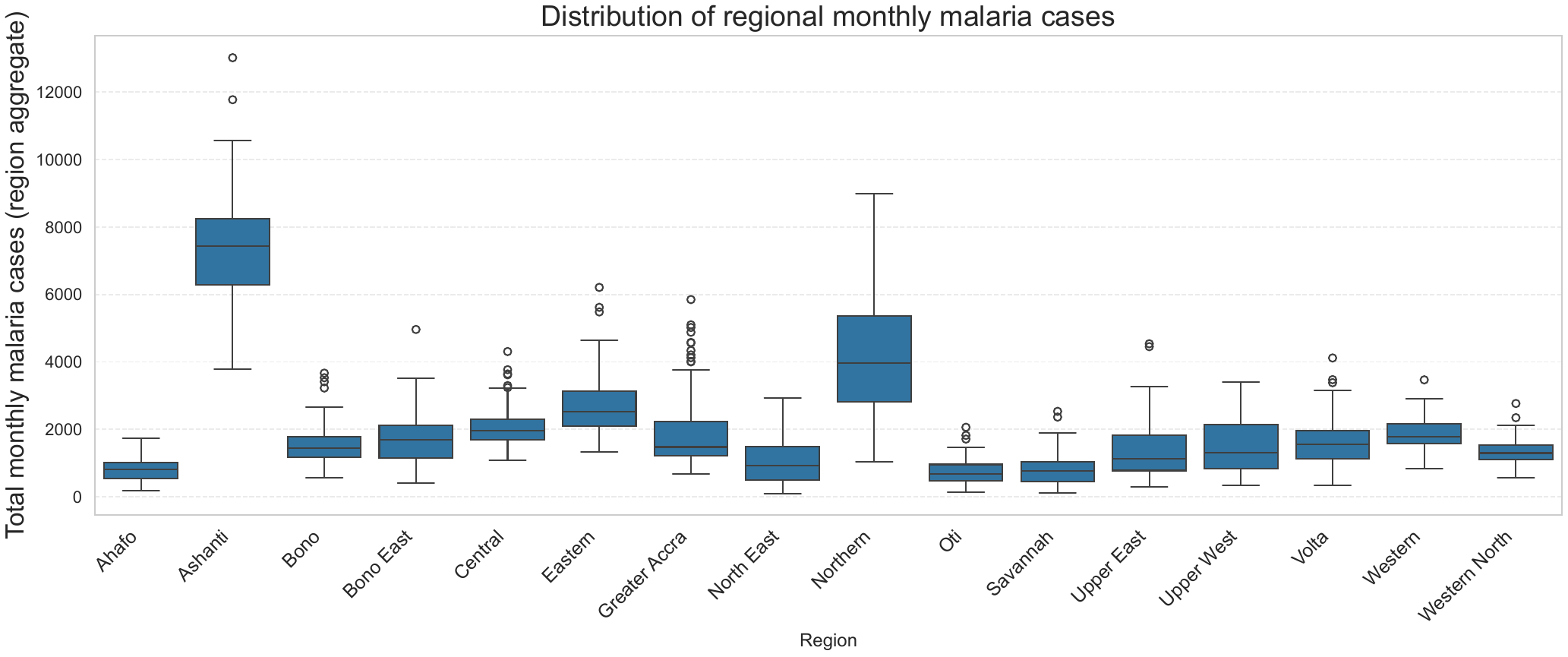} 
\end{minipage}
\caption{Distribution of monthly malaria admissions by region during 2014--2023. Boxplots summarise the median, interquartile range, overall spread, and extreme values of regional monthly case counts, illustrating marked differences in malaria burden and variability across Ghana's administrative regions.
}\label{fig:02_regional_boxplot}
\end{figure*}

Figures~\ref{fig:regional_time_series} and \ref{fig:02_regional_boxplot} provide complementary perspectives on regional malaria dynamics in Ghana during 2014--2023. The monthly time-series plots reveal substantial differences in temporal behaviour across regions, with Northern and Ashanti exhibiting persistently high malaria burdens and pronounced seasonal fluctuations, while regions such as Oti, Savannah, and Ahafo maintain comparatively lower case counts throughout the study period. Several regions, including Greater Accra and Eastern, show periods of elevated activity and changing temporal patterns, indicating that malaria transmission dynamics are not stationary across space or time. These observations are reinforced by the boxplots, which demonstrate marked differences in the distribution of monthly cases among regions. Ashanti exhibits the highest median monthly burden, whereas Northern displays the greatest variability and the widest interquartile range, reflecting strong seasonal oscillations. In contrast, regions such as Oti, Ahafo, and Savannah show narrower distributions and lower median burdens. The presence of numerous high-value outliers across multiple regions further suggests episodic periods of unusually elevated malaria activity.

Monthly malaria admission records were extracted from the DHIMS2 for the period January 2014 to December 2023. DHIMS2 has been the national platform for routine health data collation since 2012, aggregating data from all public and many private health facilities. For each district and month, the dataset contains two variables: the number of patients admitted with confirmed malaria under five years of age, and patients aged five years or older. No individual‑level clinical or laboratory data are available at the regional aggregate level. For this study, the two age groups were summed to obtain total malaria cases per district. Subsequently, cases were aggregated to the region level by summing across all districts within a region for each calendar month. This aggregation produced a region‑month panel with 16 regions over 120 months (2014--2023), yielding a theoretical maximum of 1920 observations. After constructing time‑dependent features (described in the next subsection), the complete dataset used for anomaly detection contained 1908 region‑month rows; the 12 excluded rows correspond to the first month of each region (January 2014), for which lagged variables could not be computed.

The descriptive summaries (Figs.~\ref{fig:ghmap_f1}--\ref{fig:02_regional_boxplot}) clearly identify regions such as Northern and Ashanti as having persistently high malaria burden together with pronounced seasonal variability. These analyses provide an important overview of the spatial distribution and typical temporal behaviour of malaria transmission across Ghana. However, by construction, descriptive statistics characterise the average or expected behaviour of each region and are not designed to identify observations that depart substantially from their own historical patterns. For example, an unusually sharp increase in malaria admissions during a season that normally experiences low transmission, or an unexpected decline during a period that typically records peak incidence, may represent epidemiologically important events but remain indistinguishable within conventional summaries. Similarly, a region with naturally high temporal variability, such as Northern, may experience an exceptionally extreme month that lies well beyond its historical range, whereas a relatively low-burden region, such as Oti, may exhibit a moderate increase in malaria admissions that remains far below the national average yet is highly unusual for its local transmission dynamics. In both situations, summary measures based on means, medians or interquartile ranges provide limited insight because they neither preserve the temporal ordering of observations nor account for the multivariate characteristics of malaria surveillance data, including simultaneous changes in age-specific admission patterns and seasonal behaviour.
The proposed anomaly detection framework addresses this limitation by evaluating each region-month observation relative to a region-specific epidemiological baseline that incorporates recent temporal dynamics through lag effects, expected seasonal behaviour through seasonal residuals, and within-region standardised deviations. Consequently, the framework complements rather than replaces descriptive statistical analyses. Whereas descriptive summaries characterise the overall burden and long-term transmission patterns, anomaly detection identifies statistically unusual departures from those expected patterns, thereby highlighting episodes of unexpected malaria activity that may warrant further epidemiological investigation, irrespective of whether they occur in regions with historically high or low malaria burden.

\subsection{Feature Engineering}
\label{subsec:feature_engineering}

To enable the detection of multivariate anomalies, a set of features was constructed to capture short-term dynamics, seasonal expectations, and long-term trends. Each region--month observation was described by nine variables derived from the raw surveillance data. These features were designed to reflect both the absolute level of malaria admissions and deviations from expected behaviour, thereby providing a multi-faceted representation of the transmission process.

From an epidemiological perspective, the engineered variables were selected to represent different mechanisms through which malaria transmission may depart from expected behaviour. Lag variables capture abrupt month-to-month changes that may indicate rapidly emerging transmission events. Seasonal residuals quantify deviations from the historical seasonal pattern of each region and therefore identify unusually high or unusually low transmission for a given time of year. Region-specific standardised scores enable meaningful comparison of anomaly magnitude across regions with substantially different baseline malaria burdens. Age-specific admission counts allow the framework to detect changes that disproportionately affect particular demographic groups, thereby capturing shifts in the age distribution of malaria burden.

Table~\ref{tab:features} lists each feature together with its mathematical definition and epidemiological purpose. The raw monthly total cases (\texttt{total\_cases}) serve as the baseline measure of malaria burden. To capture abrupt changes from one month to the next, the lag-one term (\texttt{lag1\_total}) was computed as the total cases in the immediately preceding month for the same region. Formally, the lag feature is defined by Eq.~\eqref{eq:lag1_total}:

\begin{equation}
\text{lag1\_total}_{r,t}
=
x_{r,t-1},
\label{eq:lag1_total}
\end{equation}

\noindent 
where \(x_{r,t-1}\) denotes the total malaria cases observed in the previous month for region \(r\).

A seasonal residual (\texttt{residual}) was obtained by subtracting the region-specific seasonal mean from the observed total cases. The seasonal mean for a given region and calendar month was defined as the average of total cases in that month across the ten years of observation. The residual feature is given by Eq.~\eqref{eq:seasonal_residual}:

\begin{equation}
\text{residual}_{r,t}
=
x_{r,t}
-
\bar{x}_{r,m},
\label{eq:seasonal_residual}
\end{equation}

\noindent
where \(x_{r,t}\) denotes the observed malaria cases for region \(r\) at time \(t\), and \(\bar{x}_{r,m}\) represents the average number of cases for region \(r\) during calendar month \(m\) computed across all years of observation.

The within-region standardised score (\texttt{region\_zscore}) was calculated using Eq.~\eqref{eq:region_zscore}:

\begin{equation}
\text{region\_zscore}
=
\frac{x-\mu_r}{\sigma_r},
\label{eq:region_zscore}
\end{equation}

\noindent
where \(x\) is the observed total cases, \(\mu_r\) is the mean malaria burden for region \(r\) over the entire study period, and \(\sigma_r\) is the corresponding regional standard deviation. This transformation normalises each region's time series to a common scale, facilitating cross-regional comparisons of anomaly strength.

Seasonal periodicity was encoded using sine and cosine transformations of the month index. The sine component was computed according to Eq.~\eqref{eq:month_sin}:

\begin{equation}
\text{month\_sin}
=
\sin\!\left(
\frac{2\pi \cdot \text{month}}{12}
\right),
\label{eq:month_sin}
\end{equation}

\noindent
while the cosine component was computed according to Eq.~\eqref{eq:month_cos}:

\begin{equation}
\text{month\_cos}
=
\cos\!\left(
\frac{2\pi \cdot \text{month}}{12}
\right),
\label{eq:month_cos}
\end{equation}

\noindent
The seasonal encoding variables defined by Eqs.~\eqref{eq:month_sin} and \eqref{eq:month_cos} capture the cyclical nature of malaria transmission without imposing a discontinuous transition between December and January. This circular representation preserves temporal proximity among months and is widely used for modelling seasonal processes.

A linear time trend (\texttt{year}) was introduced by centring the calendar year at 2014, as defined in Eq.~\eqref{eq:year_trend}:

\begin{equation}
\text{year}
=
\text{Year} - 2014,
\label{eq:year_trend}
\end{equation}



\begin{table*}[htbp]
\centering
\caption{Features constructed for anomaly detection.}
\label{tab:features}
\begin{tabular}{lp{4cm}p{6cm}}
\toprule
\textbf{Feature} & \textbf{Definition} & \textbf{Purpose} \\
\midrule

\texttt{total\_cases}
&
Raw monthly cases in region
&
Baseline malaria burden
\\

\texttt{lag1\_total}
&
Eq.~\eqref{eq:lag1_total}
&
Detect sudden month-to-month changes
\\

\texttt{residual}
&
Eq.~\eqref{eq:seasonal_residual}
&
Deviation from expected seasonality
\\

\texttt{region\_zscore}
&
Eq.~\eqref{eq:region_zscore}
&
Within-region standardisation
\\

\texttt{month\_sin}, \texttt{month\_cos}
&
Eqs.~\eqref{eq:month_sin} and \eqref{eq:month_cos}
&
Circular encoding of seasonality
\\

\texttt{year}
&
Eq.~\eqref{eq:year_trend}
&
Linear long-term trend
\\

\texttt{under5}, \texttt{over5}
&
Age-specific admission counts
&
Capture age-pattern anomalies
\\

\bottomrule
\end{tabular}
\end{table*}

\begin{figure*}[htbp]
\centering
\includegraphics[width=\textwidth]{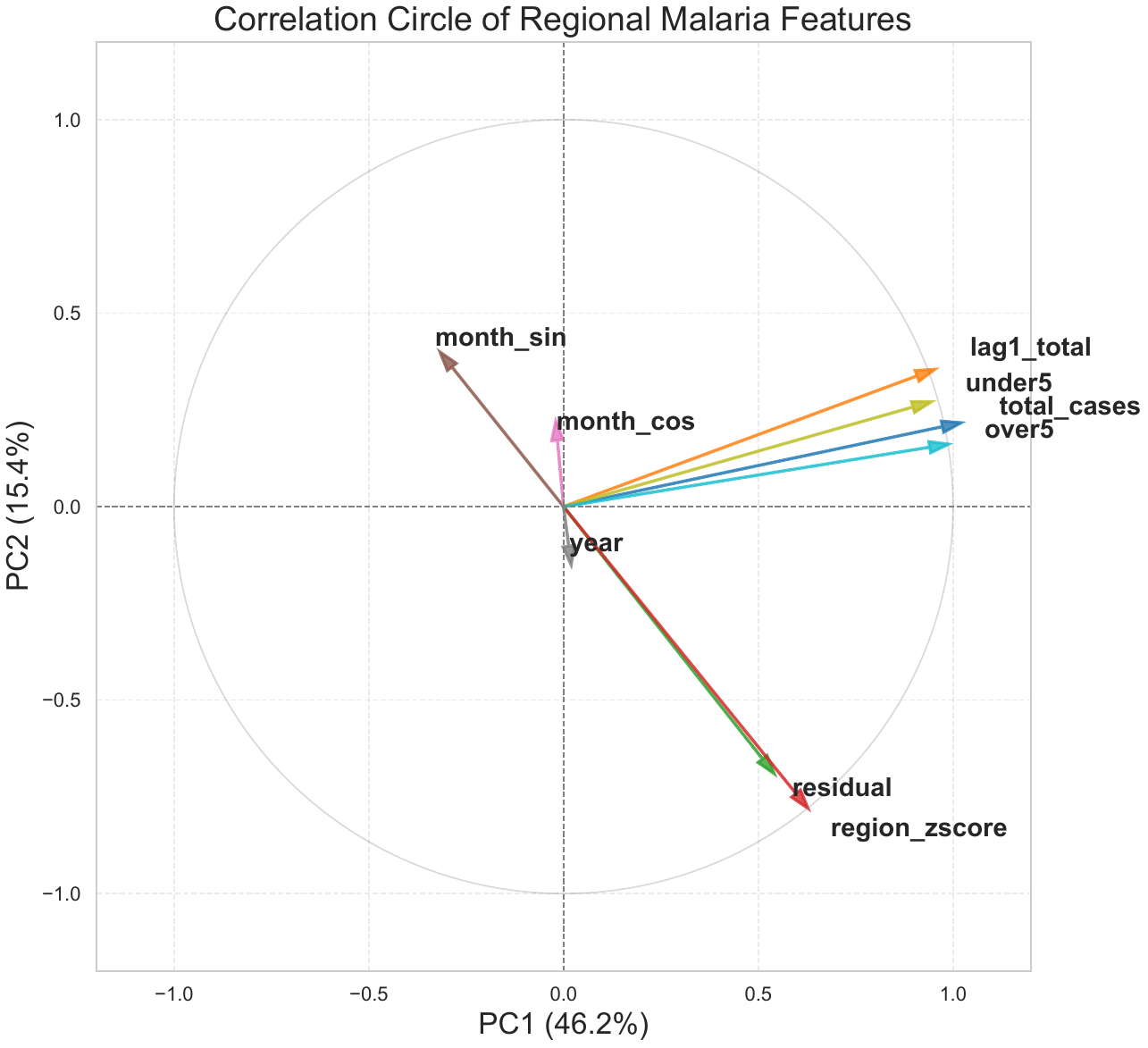}
\caption{Correlation circle from principal component analysis of the nine engineered features. The first two principal components explain $46.2\%$ and $15.4\%$ of the total variance, respectively. Features that cluster together are positively correlated, while features positioned opposite each other are negatively correlated. The \texttt{total\_cases}, \texttt{under5}, and \texttt{over5} form a tight cluster, indicating strong collinearity. \texttt{residual} and \texttt{region\_zscore} are also closely aligned, as both measure deviations from a baseline. The seasonal encodings \texttt{month\_sin} and \texttt{month\_cos} lie nearly orthogonal to the case-related features, confirming that they capture independent information about transmission timing.}
\label{fig:correlation_circle}
\end{figure*}

\noindent
thereby accounting for gradual changes in reporting practices, population growth, intervention coverage, and other long-term temporal effects over the decade.

Finally, the age-specific counts (\texttt{under5} and \texttt{over5}) were retained as separate features to allow the detection of anomalies that affect only one age group, such as outbreaks disproportionately concentrated among young children. Together, the engineered features defined by Eqs.~\eqref{eq:lag1_total}--\eqref{eq:year_trend} provide complementary information on short-term dynamics, seasonal variation, temporal trends, and demographic structure.

A principal component analysis (PCA) (refer to~\ref{app:pca}) was conducted on the standardised feature matrix to examine relationships among the engineered variables. Figure~\ref{fig:correlation_circle} presents the correlation circle for the first two principal components, which together explain $61.6\%$ of the total variance.
The analysis reveals several distinct feature groupings. The variables \texttt{total\_cases}, \texttt{under5}, \texttt{over5}, and \texttt{lag1\_total} form a tight cluster along the positive direction of the first principal component, indicating strong positive correlation among current malaria burden, age-specific admissions, and short-term temporal persistence. This pattern suggests that regions experiencing high case counts in a given month also tend to exhibit elevated counts in the preceding month.
A second cluster is formed by \texttt{residual} and \texttt{region\_zscore}, which are closely aligned with one another but separated from the primary burden variables. Their near-overlapping vectors indicate that both features quantify departures from expected transmission behaviour, albeit relative to different baselines. The strong association between these variables confirms that unusually high or low malaria activity is consistently reflected in both seasonal residuals and within-region standardised deviations.

In contrast, the seasonal encoding variables \texttt{month\_sin} and \texttt{month\_cos} occupy a different region of the feature space and are approximately orthogonal to the burden-related variables, demonstrating that seasonal timing contributes information that is largely independent of transmission magnitude. The \texttt{year} variable lies close to the origin with only a modest projection onto the first two components, suggesting that long-term temporal change contributes relatively little to the variance captured by the dominant feature dimensions.

Although several burden-related variables display substantial collinearity, the PCA indicates that the feature set captures multiple complementary aspects of malaria transmission, including burden intensity, anomaly magnitude, seasonal structure, temporal persistence, and long-term evolution. Because the anomaly detection framework relies on algorithms that are generally robust to correlated predictors, no variables were removed. Retaining the full feature set preserves information on distinct epidemiological characteristics that may contribute to anomalous behaviour.

\subsection{Anomaly Detection Algorithms} \label{subsec:anomaly_detection}

Four unsupervised anomaly detection algorithms were applied to the standardised feature matrix described in Section~\ref{subsec:feature_engineering}. Each method identifies anomalous behaviour from a distinct mathematical perspective, including random partitioning, local density estimation, non-linear reconstruction, and robust multivariate distance. Combining these complementary approaches reduces reliance on any single detection mechanism and provides a more robust basis for identifying atypical malaria transmission patterns. The outputs of the individual algorithms were subsequently integrated through a consensus framework to reduce method-specific false positives and improve the reliability of anomaly classification.

\subsubsection{Isolation Forest}

Isolation Forest (IF) identifies anomalies by recursively partitioning the feature space using random splits on randomly selected features \cite{Liu2008}. The underlying assumption is that anomalous observations are both rare and distinct, and therefore require fewer partitions to be isolated than normal observations. For an observation
\(
\mathbf{z}\in\mathbb{R}^{p}
\),
where \(p=9\) denotes the number of engineered features, an ensemble of isolation trees is constructed. Let \(h(\mathbf{z})\) denote the path length from the root node to the terminal leaf containing \(\mathbf{z}\). The anomaly score is defined by Eq.~\eqref{eq:if_score}:

\begin{equation}
s(\mathbf{z})
=
2^{-\frac{\mathbb{E}[h(\mathbf{z})]}{c(n)}},
\label{eq:if_score}
\end{equation}

\noindent where \(\mathbb{E}[h(\mathbf{z})]\) is the average path length across all trees and \(c(n)\) is the normalising factor corresponding to the expected path length of an unsuccessful search in a binary search tree. The normalisation term is given by Eq.~\eqref{eq:if_normalisation}:

\begin{equation}
c(n)
=
2H(n-1)
-
\frac{2(n-1)}{n},
\label{eq:if_normalisation}
\end{equation}

\noindent where \(H(\cdot)\) denotes the harmonic number. According to Eq.~\eqref{eq:if_score}, observations with short average path lengths receive large anomaly scores and are therefore more likely to be anomalous. In this study, 300 isolation trees were grown and the contamination parameter was fixed at 0.10.

\subsubsection{Local Outlier Factor}

Local Outlier Factor (LOF) is a density-based anomaly detection method that compares the local density surrounding an observation with the densities of its nearest neighbours \cite{Breunig2000}. Let \(N_k(\mathbf{z})\) denote the set of \(k\)-nearest neighbours of \(\mathbf{z}\), where \(k=20\). The reachability distance is defined as

\begin{equation}
\mathrm{reach\mbox{-}dist}_k
(\mathbf{z},\mathbf{z}')
=
\max
\left\{
k\text{-distance}(\mathbf{z}'),
d(\mathbf{z},\mathbf{z}')
\right\},
\label{eq:lof_reachability}
\end{equation}

\noindent  
where \(d(\cdot,\cdot)\) denotes the Euclidean distance. Using Eq.~\eqref{eq:lof_reachability}, the local reachability density (LRD) is computed as

\begin{equation}
\mathrm{lrd}_k(\mathbf{z})
=
\left(
\frac{
\sum_{\mathbf{z}'\in N_k(\mathbf{z})}
\mathrm{reach\mbox{-}dist}_k(\mathbf{z},\mathbf{z}')
}{
|N_k(\mathbf{z})|
}
\right)^{-1},
\label{eq:lof_lrd}
\end{equation}

\noindent  which measures the inverse average reachability distance around \(\mathbf{z}\). The LOF score is then obtained using Eq.~\eqref{eq:lof_score}:

\begin{equation}
\mathrm{LOF}_k(\mathbf{z})
=
\frac{1}{|N_k(\mathbf{z})|}
\sum_{\mathbf{z}'\in N_k(\mathbf{z})}
\frac{
\mathrm{lrd}_k(\mathbf{z}')
}{
\mathrm{lrd}_k(\mathbf{z})
},
\label{eq:lof_score}
\end{equation}

\noindent where values substantially greater than one indicate that the observation lies in a relatively sparse region compared with its neighbours and is therefore considered anomalous. Observations exceeding the 90th percentile of the empirical LOF distribution were labelled as anomalies.

\subsubsection{Autoencoder}

An autoencoder (AE) is  a neural network designed to learn a compressed representation of the normal data manifold and reconstruct the original input from that representation \citep{Sakurada2014}. The architecture employed in this study consisted of an input layer with 9 neurons, a bottleneck encoding layer with 4 neurons, and an output layer with 9 neurons. The network was trained by minimising the reconstruction loss defined in Eq.~\eqref{eq:ae_loss}:

\begin{equation}
\mathcal{L}_{\mathrm{AE}}
=
\frac{1}{n}
\sum_{i=1}^{n}
\left\|
\mathbf{z}^{(i)}
-
\hat{\mathbf{z}}^{(i)}
\right\|_2^2,
\label{eq:ae_loss}
\end{equation}

\noindent  where \(\hat{\mathbf{z}}^{(i)}\) denotes the reconstructed version of observation \(\mathbf{z}^{(i)}\). Training was performed using the Adam optimiser with a batch size of 32, a maximum of 150 epochs, and early stopping based on a validation split of 10\%.

Following training, the reconstruction error for each observation was computed according to Eq.~\eqref{eq:ae_error}:

\begin{equation}
e(\mathbf{z})
=
\left\|
\mathbf{z}
-
\hat{\mathbf{z}}
\right\|_2^2,
\label{eq:ae_error}
\end{equation}

\noindent where large values of Eq.~\eqref{eq:ae_error} indicate poor reconstruction and consequently anomalous behaviour. Observations with reconstruction errors above the 90th percentile threshold were classified as anomalies.

\subsubsection{Elliptic Envelope}

Elliptic Envelope (EE) assumes that the majority of observations arise from a multivariate Gaussian distribution and identifies anomalies using robust covariance estimation \cite{Rousseeuw1999}. Specifically, a Minimum Covariance Determinant (MCD) estimator is used to obtain a robust location vector \(\boldsymbol{\mu}\) and covariance matrix \(\boldsymbol{\Sigma}\).

The Mahalanobis distance of an observation is computed using Eq.~\eqref{eq:ee_mahalanobis}:

\begin{equation}
d_M(\mathbf{z})
=
\sqrt{
(\mathbf{z}-\boldsymbol{\mu})^{\top}
\boldsymbol{\Sigma}^{-1}
(\mathbf{z}-\boldsymbol{\mu})
},
\label{eq:ee_mahalanobis}
\end{equation}

\noindent which measures the distance of \(\mathbf{z}\) from the robust multivariate centre. Under the multivariate Gaussian assumption, the squared Mahalanobis distance approximately follows a chi-square distribution with \(p\) degrees of freedom. Consequently, observations satisfying Eq.~\eqref{eq:ee_threshold} were classified as anomalous:

\begin{equation}
d_M(\mathbf{z})
>
\sqrt{\chi^2_{p,0.95}},
\label{eq:ee_threshold}
\end{equation}

\noindent  where \(p=9\). A contamination parameter of 0.10 was adopted to maintain consistency across all anomaly detection methods.

\subsubsection{Consensus Rule and Anomaly Strength}

The outputs of the four anomaly detectors were combined into a consensus anomaly score. Let

\begin{equation}
\mathbb{I}_{m}
=
\begin{cases}
1, & \text{if method } m \text{ classifies the observation as anomalous},\\
0, & \text{otherwise},
\end{cases}
\label{eq:indicator_function}
\end{equation}

\noindent denote the binary anomaly indicator associated with method \(m\).

The anomaly strength was then defined according to Eq.~\eqref{eq:consensus_score}:

\begin{equation}
S
=
\mathbb{I}_{\mathrm{IF}}
+
\mathbb{I}_{\mathrm{LOF}}
+
\mathbb{I}_{\mathrm{AE}}
+
\mathbb{I}_{\mathrm{EE}},
\label{eq:consensus_score}
\end{equation}

\noindent where \(S\) represents the number of independent algorithms that classify an observation as anomalous. Rather than combining heterogeneous anomaly scores through weighted averaging, probabilistic score fusion, or meta-learning, the proposed framework first converts the output of each detector into a binary anomaly decision and subsequently aggregates these decisions using a majority-agreement rule. This strategy avoids the need to calibrate or normalise anomaly scores originating from fundamentally different detection principles while providing an interpretable measure of agreement among independent algorithms.

The rationale for adopting this consensus strategy is threefold. First, the four constituent algorithms operate according to distinct mathematical principles--namely, random partitioning (Isolation Forest), local density estimation (Local Outlier Factor), non-linear reconstruction (Autoencoder), and robust covariance estimation (Elliptic Envelope). Their anomaly scores therefore possess different numerical ranges, statistical interpretations, and calibration properties, making direct score fusion difficult without introducing additional assumptions \cite{Zimek2014, Kittler1998}. Second, the objective of the proposed framework is not to construct another composite anomaly score but rather to quantify the degree of methodological agreement supporting each detected anomaly. The consensus strength therefore provides an intuitive measure of confidence that can be readily interpreted within routine malaria surveillance. Third, requiring agreement from at least three of the four algorithms intentionally prioritises high-confidence anomaly identification by reducing algorithm-specific false positives. This represents an appropriate trade-off for operational disease surveillance, where unnecessary epidemiological investigations consume limited public health resources.

Using Eq.~\eqref{eq:consensus_score}, observations were categorised as follows:

\[
\text{Anomaly category} =
\begin{cases}
\text{Strong anomaly}, & \text{if } S = 4,\\
\text{Moderate anomaly}, & \text{if } S = 3,\\
\text{Normal}, & \text{if } S \leq 2.
\end{cases}
\]

\noindent The consensus formulation defined by Eq.~\eqref{eq:consensus_score} exploits the complementary strengths of the four anomaly detectors while reducing reliance on the assumptions of any individual method. Because each algorithm identifies anomalous behaviour from a different mathematical perspective, observations consistently identified across multiple detectors are more likely to represent robust departures from expected epidemiological behaviour than artefacts arising from the assumptions or sensitivities of a single algorithm. Consequently, conflicting outputs are resolved through the majority-agreement rule, whereby observations supported by three or four algorithms are retained as consensus anomalies, while those identified by fewer than three methods are regarded as normal observations. Although the consensus threshold is user-configurable, lowering the threshold (\(S \geq 2\)) would increase sensitivity at the expense of specificity, whereas the threshold adopted in this study (\(S \geq 3\)) was selected to prioritise high-confidence surveillance signals suitable for public health decision support.

The entire anomaly detection pipeline, from feature standardisation to consensus classification, was implemented in Python using \texttt{scikit-learn} (v1.2) \cite{pedregosa2011scikit} for IF, LOF, and EE, and \texttt{TensorFlow}\footnote{\url{https://www.tensorflow.org/}} (v2.10) for the AE. Random seeds were fixed throughout to ensure reproducibility. For the complete dataset (\(n = 1908\) observations), all four models were trained in under 30 seconds on a standard workstation (Intel Core i7 processor with 16\,GB RAM), demonstrating the practical feasibility of the proposed framework for routine malaria surveillance applications.

\subsection{Evaluation and Spatial Interpolation Metrics}
\label{subsec:evaluation_metrics}

Malaria surveillance data are often characterised by non-normal distributions, spatial sparsity, temporal dependence, and occasional extreme outbreaks. Consequently, conventional parametric procedures that assume normality, such as the \(t\)-test, ANOVA, and ordinary least-squares regression, may not adequately capture the complexity of epidemiological observations. Likewise, simple threshold-based approaches that compare observed case counts against historical averages are limited because they ignore temporal ordering, fail to exploit multivariate information, and are insensitive to subtle but epidemiologically meaningful departures from expected behaviour.

To evaluate the consensus anomaly framework and investigate the spatial manifestation of anomalous transmission patterns, a complementary set of statistical and visualisation metrics was employed. These metrics quantify distributional differences, practical significance, inter-method agreement, latent feature-space structure, and spatial continuity while respecting the underlying characteristics of the surveillance data.

\subsubsection{Mann--Whitney U Test for Distributional Differences}

The first stage of evaluation examined whether anomalous months differ systematically from normal months with respect to the engineered features. Because the feature distributions are generally skewed and may contain extreme observations, the non-parametric Mann--Whitney U test (Wilcoxon rank-sum test) was selected.

Let

\[
X=\{x_1,\ldots,x_{n_X}\}
\]

\noindent denote observations corresponding to anomalous months (\(S\ge3\)) and

\[
Y=\{y_1,\ldots,y_{n_Y}\}
\]

\noindent  denote observations corresponding to normal months (\(S\le2\)). The Mann--Whitney statistic is defined as

\begin{equation}
U
=
\sum_{i=1}^{n_X}
\sum_{j=1}^{n_Y}
\mathbf{1}(x_i>y_j),
\label{eq:mann_whitney}
\end{equation}

\noindent  where \(\mathbf{1}(\cdot)\) denotes the indicator function.

Equation~\eqref{eq:mann_whitney} counts the number of pairwise comparisons in which an anomalous observation exceeds a normal observation. Under the null hypothesis, both samples originate from the same distribution. A two-sided hypothesis test was performed and statistical significance was declared when \(p<0.05\). The Mann--Whitney test therefore addresses the question of whether anomalous months exhibit systematically different feature distributions compared with normal months.

\subsubsection{Cohen's \texorpdfstring{$d$}{d} for Effect Size}

While the Mann--Whitney test identifies statistically significant differences, significance alone does not quantify the magnitude of those differences. In large epidemiological datasets, small differences may become statistically significant even when they have limited practical relevance. Therefore, Cohen's \(d\) was computed to measure effect size.

The standardised mean difference is defined as

\begin{equation}
d
=
\frac{\bar{X}-\bar{Y}}
{s_p},
\label{eq:cohen_d}
\end{equation}

\noindent where \(s_p\) is the pooled standard deviation,

\begin{equation}
s_p
=
\sqrt{
\frac{
(n_X-1)s_X^2
+
(n_Y-1)s_Y^2
}
{n_X+n_Y-2}
},
\label{eq:pooled_sd}
\end{equation}

\noindent  and \(\bar{X}\), \(\bar{Y}\), \(s_X^2\), and \(s_Y^2\) denote the sample means and variances of the anomalous and normal groups, respectively.

\noindent Substituting Eq.~\eqref{eq:pooled_sd} into Eq.~\eqref{eq:cohen_d} yields a scale-independent measure of practical significance. Effect sizes were interpreted according to conventional thresholds: small (\(|d|\ge0.2\)), medium (\(|d|\ge0.5\)), and large (\(|d|\ge0.8\)). Reporting Cohen's \(d\) alongside the Mann--Whitney test provides a complementary assessment of both statistical and epidemiological significance.

\subsubsection{Cohen's Kappa for Inter-Method Agreement}

To quantify the pairwise agreement among the four anomaly detection outputs, Cohen’s Kappa coefficient was computed for each method pair as

\begin{equation}
\kappa
=
\frac{p_o-p_e}
{1-p_e},
\label{eq:kappa}
\end{equation}

\noindent where \(p_o\) denotes the observed proportion of agreement and \(p_e\) represents the agreement expected by chance under independent classification.
Equation~\eqref{eq:kappa} adjusts raw agreement for random coincidence and therefore provides a more meaningful measure of classifier consistency. Values of \(\kappa\) close to unity indicate strong agreement beyond chance, whereas values near zero suggest agreement no better than random assignment.

Interpretation followed the Landis--Koch scale: slight (0.01--0.20), fair (0.21--0.40), moderate (0.41--0.60), substantial (0.61--0.80), and almost perfect (0.81--0.99). This analysis helps identify both complementary methods (low \(\kappa\)) and potentially redundant methods (high \(\kappa\)) within the ensemble framework.

\subsubsection{UMAP for Dimensionality Reduction and Visualisation}

After validating the anomaly classifications statistically, the next step was to investigate whether anomalous observations occupy distinct regions within the multivariate feature space.
Although PCA provides a useful linear representation, it may fail to preserve non-linear structures that frequently arise in epidemiological datasets. Therefore, Uniform Manifold Approximation and Projection (UMAP) was employed to generate a two-dimensional embedding of the nine-dimensional feature space.

UMAP constructs a weighted nearest-neighbour graph in the original feature space and seeks a low-dimensional representation that preserves both local and global topological structure. Specifically, the algorithm minimises the cross-entropy objective

\begin{equation}
\mathcal{L}
=
\sum_{(i,j)}
\Big[
p_{ij}\log\!\left(\frac{p_{ij}}{q_{ij}}\right)
+
(1-p_{ij})
\log\!\left(
\frac{1-p_{ij}}
{1-q_{ij}}
\right)
\Big],
\label{eq:umap_loss}
\end{equation}

\noindent where \(p_{ij}\) and \(q_{ij}\) represent pairwise neighbourhood probabilities in the high-dimensional and low-dimensional spaces, respectively.
The embedding was generated using \(n_{\text{neighbors}}=15\) and \(\text{min\_dist}=0.1\). Equation~\eqref{eq:umap_loss} encourages the low-dimensional representation to preserve the manifold structure of the original feature space. 

\subsubsection{RBF Spatial Interpolation for Spatial}

The anomaly detection framework described in Section~\ref{subsec:anomaly_detection} identifies statistically unusual region--month observations using a multivariate representation of malaria transmission dynamics. Each observation is characterised by engineered epidemiological features that capture recent temporal behaviour, seasonal departures, long-term trends, and age-specific admission patterns. Consequently, anomaly detection is performed relative to the historical behaviour of each region rather than relative to neighbouring regions.

This design deliberately separates the objectives of anomaly detection and spatial interpretation. The primary objective of the consensus framework is to determine whether the epidemiological characteristics of a given region--month observation deviate substantially from its expected historical behaviour. Explicit modelling of spatial autocorrelation through neighbourhood structures or spatial weighting matrices was therefore not incorporated into the detection algorithms. Such extensions require additional assumptions regarding neighbourhood definitions, spatial weights, or distance-decay relationships, which lie beyond the scope of the present study and are not natively supported by the unsupervised algorithms employed.

Once consensus anomalies have been identified, their geographical distribution can then be examined to determine whether anomalous observations exhibit coherent spatial organisation. For this purpose, Radial Basis Function (RBF) interpolation was employed to transform discrete district observations into continuous spatial surfaces. This interpolation does not contribute to anomaly identification or anomaly strength estimation. Rather, it provides a continuous representation of the geographical distribution of malaria burden and consensus anomaly intensity, facilitating interpretation of regional gradients, potential hotspot structures, and broader spatial patterns across Ghana.

Let $(x_i,y_i)$ denote the coordinates of district centroid $i$, and let $z_i$ represent the corresponding malaria burden. The interpolated surface is defined by

\begin{equation}
\hat{z}(x,y)
=
\sum_{i=1}^{N}
w_i
\,
\phi
\!\left(
\sqrt{
(x-x_i)^2+(y-y_i)^2
}
\right),
\label{eq:rbf_interpolation}
\end{equation}

\noindent where \(w_i\) are interpolation weights and \(\phi(\cdot)\) is the radial basis kernel.

A multiquadric basis function was adopted,

\begin{equation}
\phi(r)
=
\sqrt{r^2+\varepsilon^2},
\label{eq:multiquadric_kernel}
\end{equation}

\noindent with smoothing parameter \(\varepsilon=0.5\).
The interpolation weights were obtained by solving

\begin{equation}
\mathbf{\Phi}\mathbf{w}
=
\mathbf{z},
\label{eq:rbf_weights}
\end{equation}

\noindent  where the kernel matrix satisfies

\begin{equation}
\Phi_{ij}
=
\phi
\!\left(
\left\|
(x_i,y_i)
-
(x_j,y_j)
\right\|
\right).
\label{eq:rbf_kernel_matrix}
\end{equation}

Equations~\eqref{eq:rbf_interpolation}--\eqref{eq:rbf_kernel_matrix} define the continuous interpolation surface used for mapping. The resulting grids were clipped to administrative boundaries and constrained to non-negative values. Importantly, these surfaces were used solely for descriptive visualisation and hypothesis generation. No inferential, predictive, or causal interpretation was attached to the interpolated outputs.

\section{Results and Analysis}\label{sec:RNA}

\subsection{Spatiotemporal Patterns of Regional Anomalies}

\begin{figure*}
\begin{minipage}[H]{\linewidth}
\centering
\includegraphics[width=\textwidth]{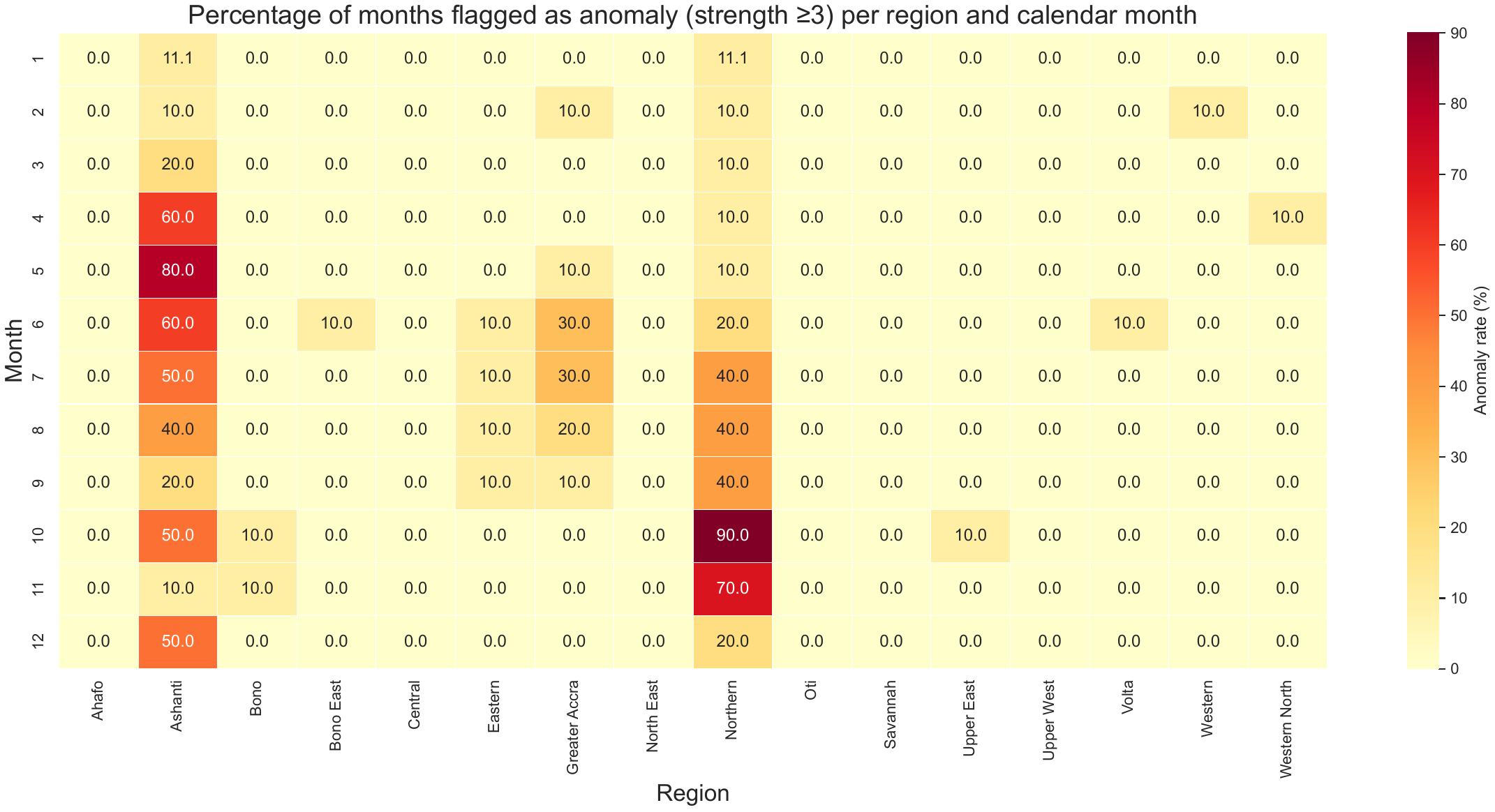} 
\end{minipage}
\caption{Percentage of region--month observations classified as consensus anomalies ($S \geq 3$) across the 16 administrative regions of Ghana during 2014--2023. Values represent the proportion of years in which a given calendar month was flagged as anomalous by at least three of the four unsupervised detection algorithms. Warmer colours indicate higher anomaly frequencies. The figure highlights pronounced seasonal concentration of anomalies in Ashanti and Northern Regions, with comparatively limited anomaly occurrence across most other regions.
}\label{fig:seasonal_anomaly_heatmap}
\end{figure*}

\begin{figure*}
\begin{minipage}[H]{\linewidth}
\centering
\includegraphics[width=\textwidth]{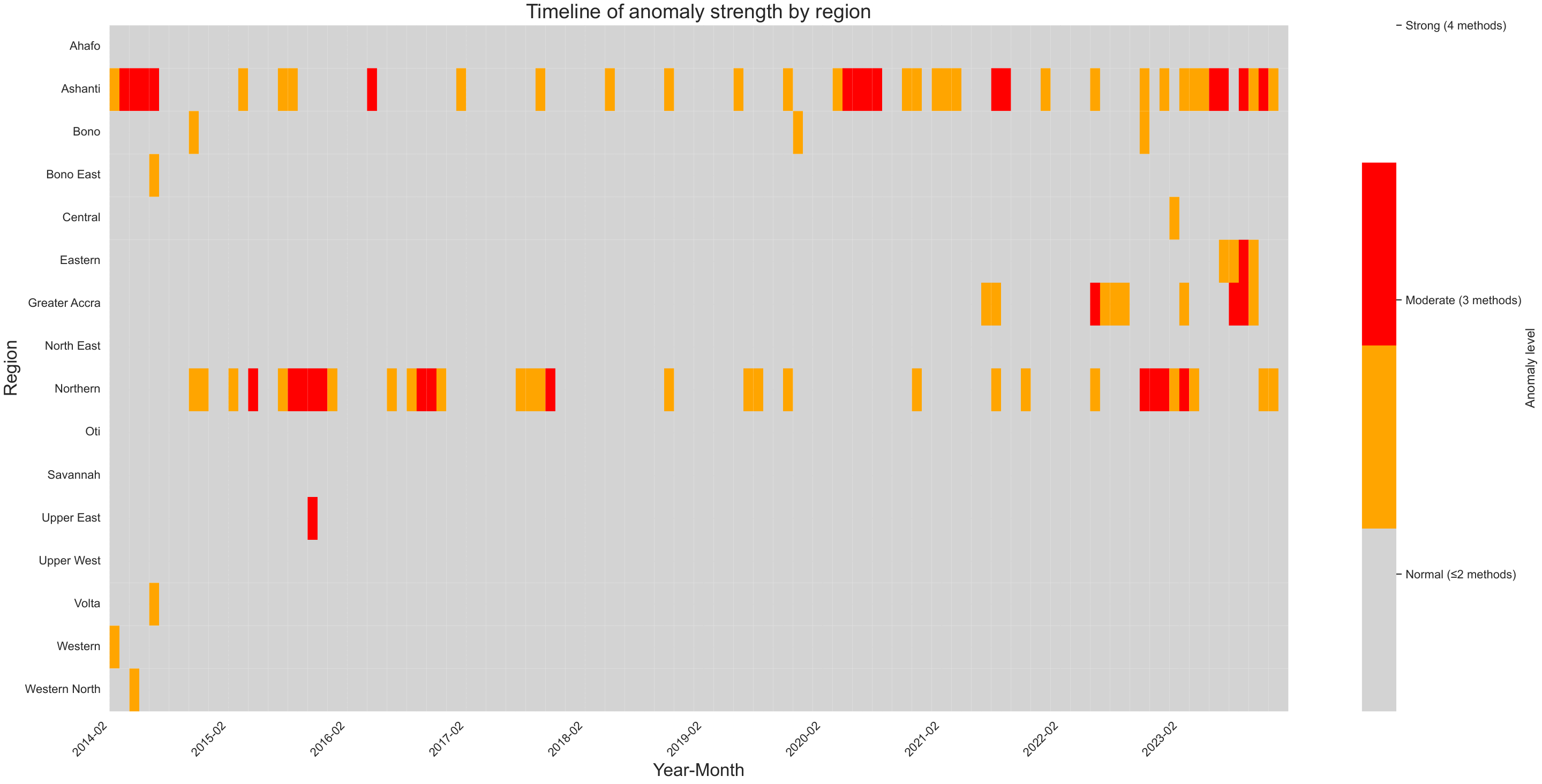} 
\end{minipage}
\caption{Temporal distribution of consensus anomaly strength across Ghana's 16 administrative regions from January 2014 to December 2023. Each cell represents a region--month observation classified according to the number of anomaly detection algorithms in agreement. Normal observations correspond to agreement by two or fewer methods ($S \leq 2$), moderate anomalies correspond to agreement by three methods ($S = 3$), and strong anomalies correspond to agreement by all four methods ($S = 4$). The figure provides a chronological view of the occurrence, persistence, and clustering of anomalous malaria activity throughout the study period.
}\label{fig:anomaly_timeline_heatmap}
\end{figure*}

\begin{figure*}
\begin{minipage}[H]{\linewidth}
\centering
\includegraphics[width=\textwidth]{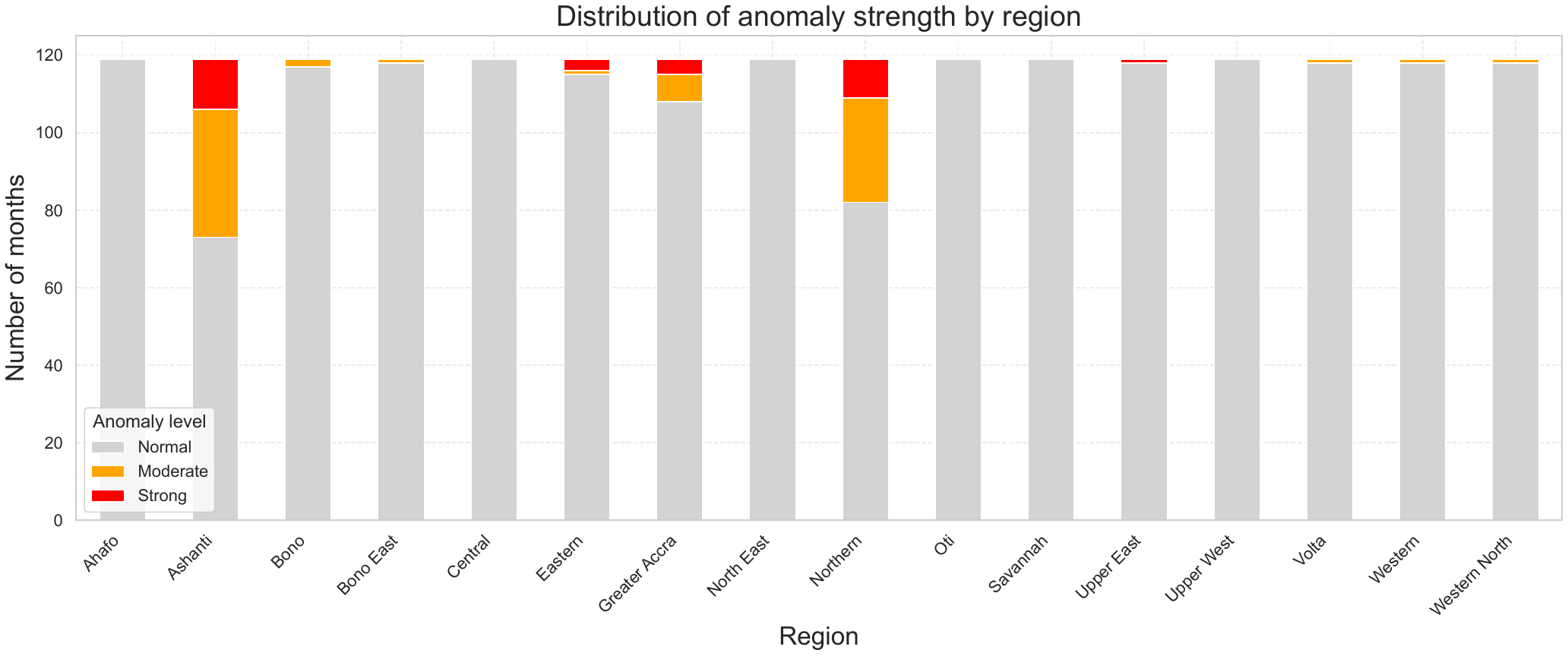} 
\end{minipage}
\caption{Distribution of anomaly classifications by region during 2014--2023. Stacked bars show the total number of region--month observations classified as normal ($S \leq 2$), moderate anomaly ($S = 3$), or strong anomaly ($S = 4$) according to the consensus framework. The figure summarises the relative contribution of each region to the overall anomaly burden and illustrates substantial regional differences in the frequency of anomalous malaria activity.
}\label{fig:anomaly_counts_region}
\end{figure*}

Figures~\ref{fig:seasonal_anomaly_heatmap}, \ref{fig:anomaly_timeline_heatmap}, and \ref{fig:anomaly_counts_region} summarise the spatial and temporal distribution of consensus anomalies identified across Ghana between 2014 and 2023. The results reveal substantial heterogeneity in both the frequency and timing of anomalous malaria activity, indicating that deviations from expected transmission patterns were concentrated within a small subset of regions rather than being uniformly distributed across the country.

The seasonal anomaly heatmap (Fig.~\ref{fig:seasonal_anomaly_heatmap}) demonstrates that anomaly occurrence was strongly structured by calendar month and region. Ashanti and Northern Regions exhibited the highest anomaly frequencies throughout the study period, with several months recording anomaly rates exceeding $40\%$ of observations. In Ashanti, elevated anomaly frequencies were concentrated between April and August, reaching a maximum of $80\%$ in May. Northern Region displayed a broader seasonal distribution, with anomalies occurring across much of the year and peaking at $90\%$ in October. By contrast, most other regions exhibited either isolated anomalous months or no recurrent seasonal anomaly signature. Greater Accra showed moderate anomaly activity during the middle of the year, particularly from June to August, while the Eastern Region recorded lower but persistent anomaly occurrence during the same seasonal window. Several regions, including Ahafo, Central, Oti, Savannah, and Upper West, exhibited no recurring seasonal anomaly pattern, with anomaly frequencies remaining close to zero across all calendar months. These results indicate that anomalous malaria activity was concentrated within specific ecological and epidemiological settings rather than occurring synchronously across Ghana \cite{Aheto2022, Kigozi2020}.

Temporal evolution of anomaly strength is presented in Fig.~\ref{fig:anomaly_timeline_heatmap}. Distinct clusters of moderate and strong anomalies were observed in a limited number of regions, with Ashanti and Northern Regions accounting for the majority of recurrent events. Ashanti experienced repeated episodes throughout the decade, including clusters during 2014, 2015, 2020, 2021, and 2023. Northern Region displayed a similarly persistent pattern, with notable concentrations during 2015--2017 and renewed activity between 2022 and 2023. Greater Accra exhibited comparatively few anomalies during the early years of the record but showed increased activity after 2021. Eastern Region followed a similar pattern, with anomaly events becoming more apparent towards the end of the study period. In contrast, several regions recorded only isolated anomaly occurrences, often confined to a single month or year. The temporal distribution therefore suggests that anomalous behaviour was not associated with a single nationwide event but instead reflected region-specific departures from expected transmission dynamics occurring at different times across the surveillance period.

Additional evidence is provided by the regional anomaly counts shown in Fig.~\ref{fig:anomaly_counts_region}. The majority of region-month observations were classified as normal, indicating that agreement among at least three anomaly detection algorithms was relatively uncommon. Nevertheless, marked regional differences emerged in the distribution of moderate and strong anomalies. Ashanti recorded the largest number of anomalous months, followed by Northern Region. Both regions contained substantial numbers of observations classified as moderate anomalies, together with multiple strong anomalies identified by all four detection algorithms. Greater Accra and Eastern Regions also contributed a measurable number of anomalous observations, although at substantially lower frequencies. Most remaining regions recorded only one or two anomaly months during the entire study period, while several contained no strong anomalies. The concentration of anomaly counts within a small number of regions indicates that consensus departures from expected malaria behaviour were geographically uneven and were dominated by a limited set of transmission environments.

Comparison of Figs.~\ref{fig:seasonal_anomaly_heatmap} and \ref{fig:anomaly_counts_region} further shows that regions with the highest anomaly burden were not necessarily those exhibiting anomalies in all seasons. Instead, recurrent anomalies tended to occur within specific temporal windows. For example, Ashanti displayed pronounced anomaly activity during the principal rainy season months, whereas Northern Region exhibited elevated anomaly frequencies extending into the later months of the year. This distinction is consistent with the differing rainfall regimes that characterise Ghana's forest and savannah ecological zones, where bimodal rainfall patterns dominate in the south and centre, while a unimodal rainy season governs much of northern Ghana \cite{OhenebaDornyo2022, Asare2017}. The anomaly framework therefore, captured departures from region-specific seasonal expectations rather than merely identifying periods of high malaria burden.

As a whole, the three visualisations indicate that anomalous malaria activity during 2014--2023 was concentrated primarily within Ashanti and Northern Regions, with secondary contributions from Greater Accra and Eastern Regions. The anomalies occurred intermittently through time and displayed clear seasonal preferences, supporting the presence of geographically distinct transmission dynamics across Ghana's ecological zones.

\subsection{Characterisation of High-Confidence Malaria Anomalies}

\begin{figure*}
\begin{minipage}[H]{\linewidth}
\centering
\includegraphics[width=\textwidth]{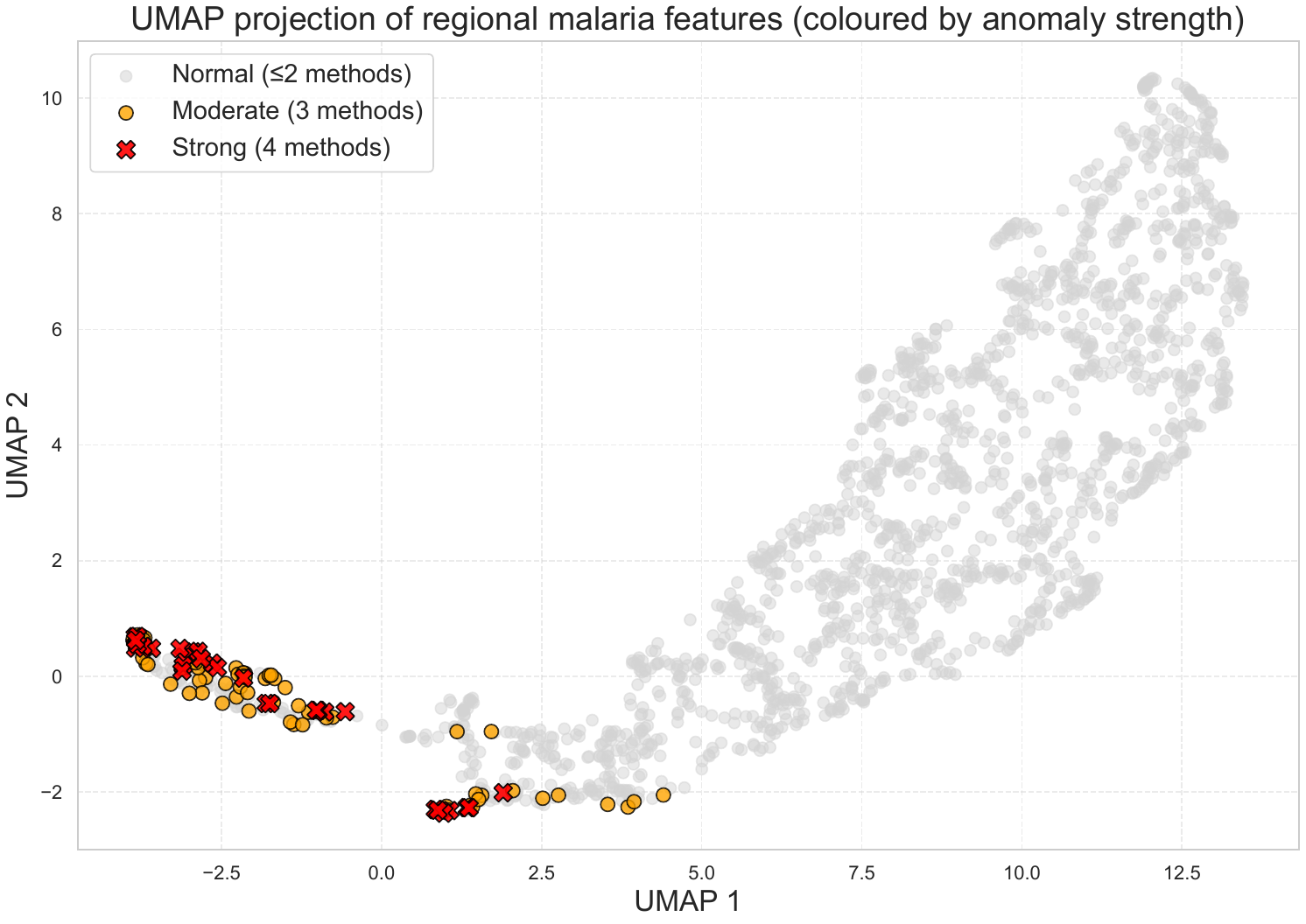} 
\end{minipage}
\caption{Two-dimensional UMAP embedding of the 1,908 region--month observations constructed from the nine engineered malaria surveillance features. Each point represents a single region--month observation and is coloured according to its consensus anomaly classification: normal ($S \leq 2$), moderate anomaly ($S = 3$), and strong anomaly ($S = 4$), where $S$ denotes the number of anomaly detection algorithms in agreement. The projection preserves local neighbourhood structure in the high-dimensional feature space, allowing visual assessment of the separation between normal and anomalous malaria transmission patterns.
}\label{fig:umap_anomalies}
\end{figure*}

The consensus framework identified a relatively small subset of region--month observations that were consistently classified as anomalous by multiple algorithms. Examination of these events in the reduced feature space revealed a clear separation between normal observations and high-confidence anomalies. Figure~\ref{fig:umap_anomalies} shows the two-dimensional UMAP embedding of the nine engineered features. Most observations formed a continuous and densely populated manifold representing the dominant patterns of malaria transmission across Ghana. In contrast, moderate and strong anomalies occupied peripheral regions of the embedding, with strong anomalies concentrated within a compact and largely distinct cluster. This separation indicates that the strongest anomaly events differed simultaneously across several feature dimensions rather than being driven by a single extreme variable. The result is consistent with the purpose of UMAP, which preserves local neighbourhood structure while capturing broader relationships within high-dimensional datasets \citep{2018arXiv180203426M}.

\begin{figure*}
\begin{minipage}[H]{\linewidth}
\centering
\includegraphics[width=\textwidth]{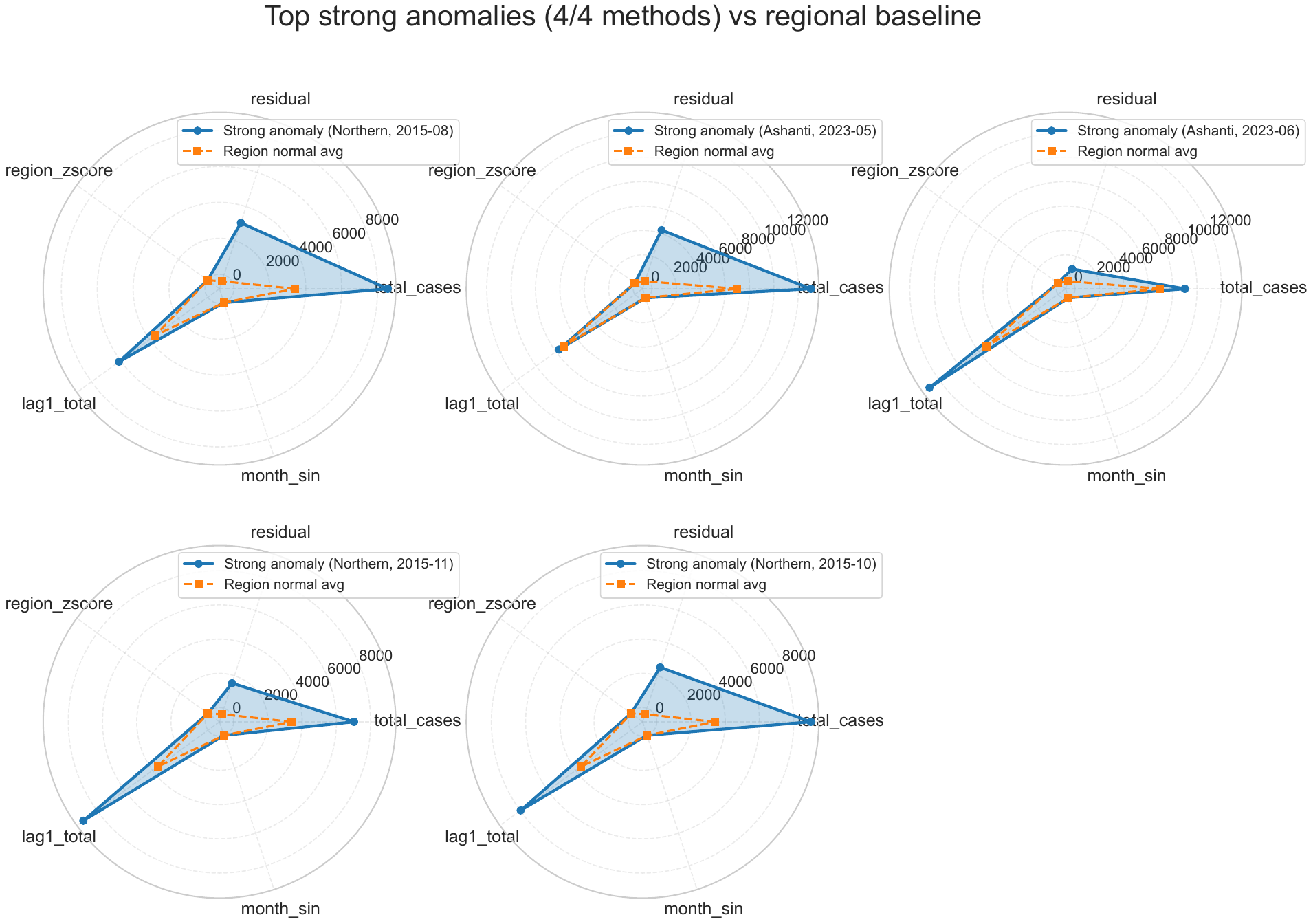} 
\end{minipage}
\caption{Radar-chart comparison of the five strongest anomaly events identified by the consensus framework. Each panel contrasts the feature profile of a strong anomaly detected by all four anomaly detection algorithms with the corresponding long-term regional baseline. Variables include total malaria cases, lagged malaria burden, seasonal residual, within-region standardised anomaly score, and seasonal phase. The figure illustrates the multivariate characteristics that distinguish high-confidence anomaly events from typical transmission conditions within the same region.
}\label{fig:radar_top_anomalies}
\end{figure*}

Inspection of the strongest anomaly events revealed that they were concentrated within a limited number of regions and years. Northern Region accounted for several of the highest-ranked anomalies, particularly during 2015, while Ashanti Region contributed the most prominent recent events during 2023 (Fig.~\ref{fig:radar_top_anomalies}). These observations correspond to months that were classified as anomalous by all four detection algorithms and therefore represent the highest level of consensus within the ensemble framework. Their concentration within a small number of regions is consistent with the spatial patterns identified earlier, where recurrent anomalies were disproportionately associated with regions already exhibiting substantial temporal variability.

Radar profiles of the top five strong anomalies demonstrate that the dominant distinguishing characteristics were elevated total malaria burden, large positive seasonal residuals, and pronounced within-region standardised deviations from historical expectations. Northern Region during August, October, and November 2015 exhibited substantially higher malaria admissions than the corresponding regional baseline, accompanied by marked increases in residual magnitude and anomaly scores. Similar behaviour was observed in the Ashanti Region during May and June 2023, where observed malaria burden exceeded typical regional conditions across multiple feature dimensions. Notably, lagged case counts were also elevated in several of these events, indicating that strong anomalies frequently occurred during periods of sustained transmission rather than as isolated monthly spikes. The combination of high case burden and large positive residuals suggests that these events reflected departures from expected seasonal behaviour rather than simple manifestations of regular seasonal peaks.

Another notable feature of the strongest anomalies is the simultaneous elevation of both absolute and relative measures of transmission. Total case counts quantify the magnitude of malaria burden, whereas the residual and region-specific standardised score capture deviation from expected local conditions. Several strong anomalies exhibited large departures in both measures, indicating that the identified events were not merely associated with historically high-burden regions but also represented unusually elevated transmission relative to each region's own climatological baseline. Such departures are particularly relevant in endemic settings where seasonal variability is expected and where anomaly detection must distinguish between normal seasonal peaks and genuinely unusual transmission episodes \cite{nakakana2020validation, Gething2010}.

\begin{figure*}
\begin{minipage}[H]{\linewidth}
\centering
\includegraphics[width=\textwidth]{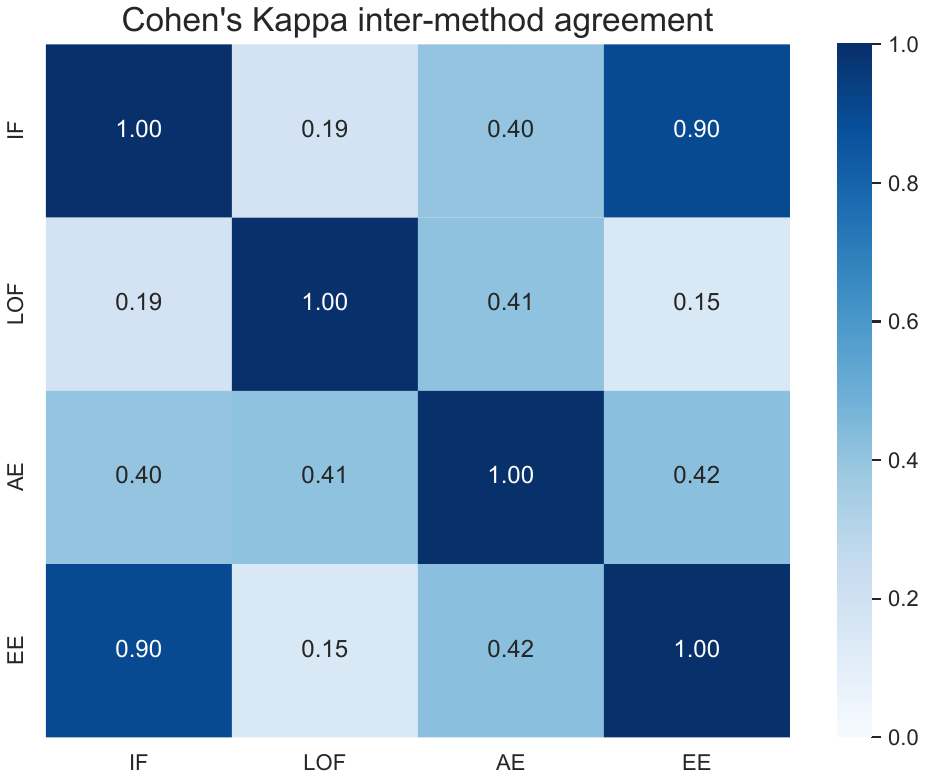} 
\end{minipage}
\caption{Pairwise Cohen's Kappa coefficients measuring agreement among the four anomaly detection algorithms: Isolation Forest (IF), Local Outlier Factor (LOF), Autoencoder (AE), and Elliptic Envelope (EE). Kappa values quantify agreement beyond chance, with larger values indicating stronger consistency in anomaly classification. The matrix provides an assessment of methodological concordance and complementarity within the consensus anomaly detection framework.
}\label{fig:kappa_matrix}
\end{figure*}

The robustness of these classifications can be assessed through the inter-method agreement matrix shown in Fig.~\ref{fig:kappa_matrix}. Agreement levels varied considerably across the four algorithms, reflecting their differing mathematical assumptions and sensitivity to distinct forms of anomalous behaviour. The strongest agreement was observed between Isolation Forest and Elliptic Envelope ($\kappa = 0.90$), indicating that both methods frequently identified the same observations as anomalous. Agreement between Autoencoder and Elliptic Envelope was moderate ($\kappa = 0.47$), while Isolation Forest and Autoencoder also showed moderate correspondence ($\kappa = 0.42$). In contrast, Local Outlier Factor exhibited comparatively weak agreement with the remaining methods, with pairwise coefficients ranging from 0.15 to 0.31.

These differences are informative because each algorithm evaluates abnormality from a different perspective. Isolation Forest isolates observations through recursive partitioning, Elliptic Envelope relies on robust multivariate distance, Autoencoders identify departures from learned feature representations, and Local Outlier Factor assesses local density structure. The relatively low agreement involving Local Outlier Factor suggests that local neighbourhood irregularities do not always coincide with globally unusual observations. At the same time, the moderate-to-high agreement among the remaining methods indicates the presence of a stable core set of anomaly events that can be detected using different analytical principles. This pattern supports the use of a consensus strategy, as reliance on a single algorithm would either overlook some anomalous observations or increase susceptibility to method-specific artefacts \cite{Aggarwal2017, Chandola2009}.

The separation observed in the UMAP embedding, together with the multivariate departures illustrated by the radar profiles and the agreement structure of the anomaly detectors, demonstrates that the strongest anomaly events represent coherent departures from regional transmission norms rather than isolated statistical outliers. These observations therefore constitute suitable candidates for subsequent epidemiological investigation and spatial analysis within the national malaria surveillance system.

\subsection{District-Level Hotspots Underlying Regional Anomaly Patterns}

\begin{figure*}
\begin{minipage}[H]{\linewidth}
\centering
\includegraphics[width=\textwidth]{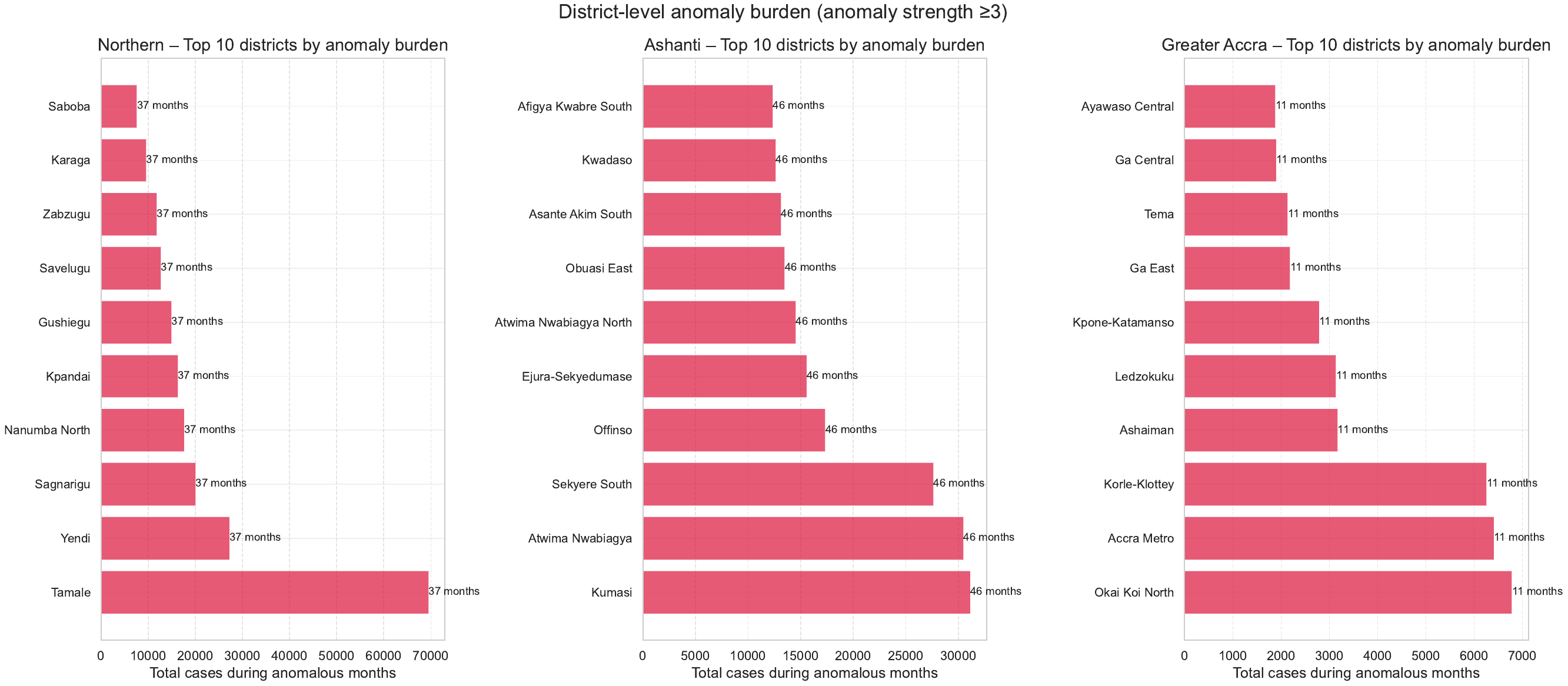} 
\end{minipage}
\caption{District-level malaria burden during anomalous months within Northern, Ashanti, and Greater Accra Regions. Bars represent cumulative malaria admissions recorded during months classified as anomalous by at least three anomaly detection methods ($S \geq 3$). Labels indicate the total number of anomalous months experienced by each district between 2014 and 2023. Districts are ranked according to cumulative anomaly-associated malaria burden.
}\label{fig:district_bar_chart}
\end{figure*}

Regional anomaly patterns were driven by a relatively small number of districts that repeatedly contributed disproportionate malaria burdens during anomalous periods. Figure~\ref{fig:district_bar_chart} reveals marked concentration of anomaly-associated malaria cases within Northern, Ashanti, and Greater Accra Regions. The strongest concentration occurred in Northern Region, where Tamale recorded approximately 70,000 malaria cases during anomalous months, substantially exceeding all other districts in the region. Yendi formed a secondary hotspot with roughly 27,000 cases, while the remaining districts each contributed fewer than 20,000 cases. Notably, all ten districts experienced the same number of anomalous months (37 months), indicating that differences in cumulative burden were driven primarily by the magnitude of malaria transmission during anomalous periods rather than by differences in anomaly frequency.

Ashanti Region displayed a different structure. Kumasi, Atwima Nwabiagya, and Sekyere South each accumulated close to 30,000 malaria cases during anomalous months, while Offinso, Ejura-Sekyedumase, Atwima Nwabiagya North, and Obuasi East formed a second tier of districts with burdens ranging between approximately 14,000 and 18,000 cases. All ten districts experienced 46 anomalous months, suggesting a highly synchronised anomaly pattern across the region. Compared with Northern Region, anomaly burden was distributed across a broader group of districts rather than being dominated by a single urban centre. Greater Accra exhibited substantially lower malaria burdens, with the leading districts, including Okai Koi North, Accra Metropolitan, and Korle-Klottey, each contributing between 6,000 and 7,000 cases during anomalous months. The uniform occurrence of 11 anomalous months across all ten districts again indicates that variation in cumulative burden largely reflected differences in case volume rather than anomaly persistence.

\begin{figure*}
\begin{minipage}[H]{\linewidth}
\centering
\includegraphics[width=\textwidth]{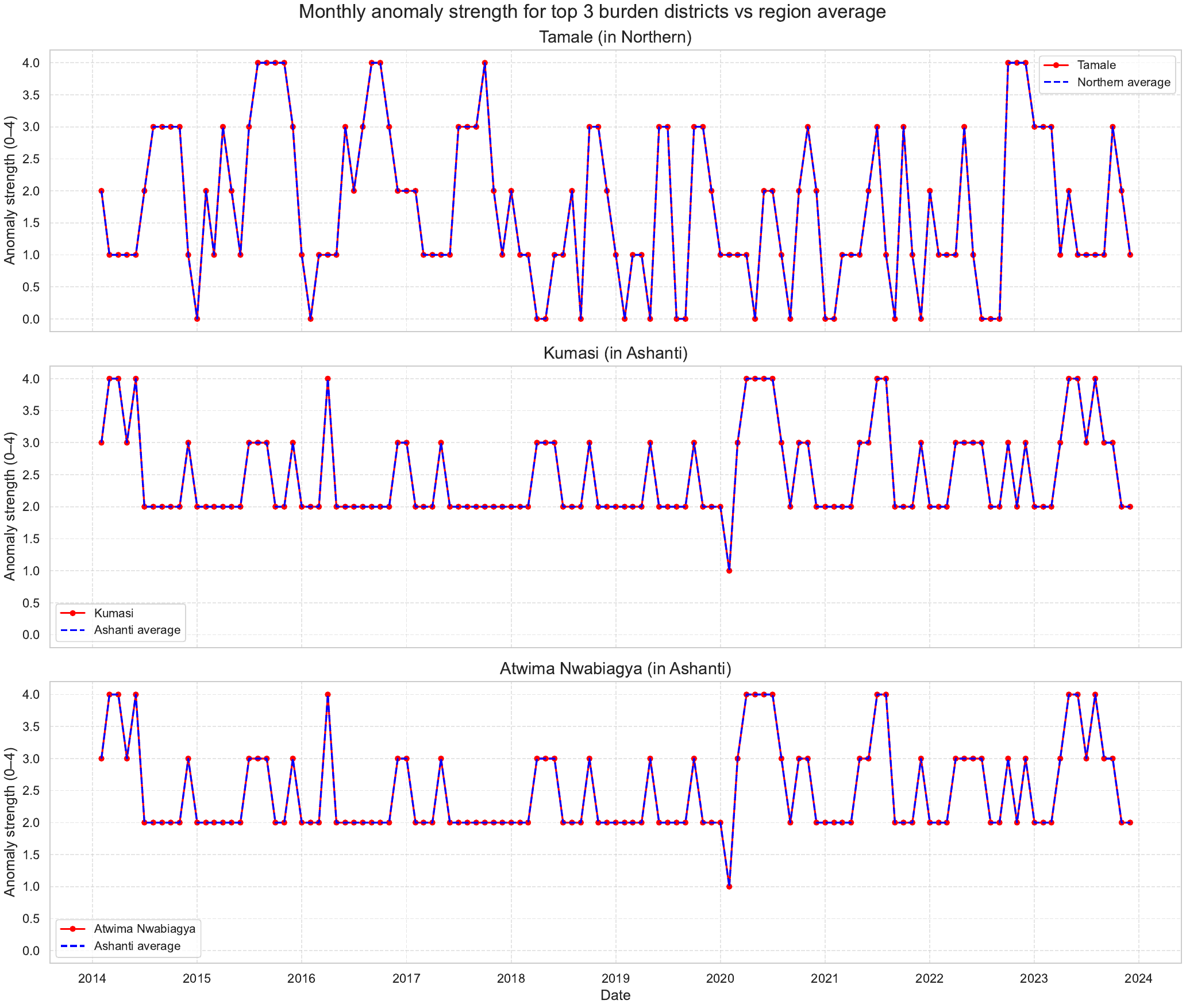} 
\end{minipage}
\caption{Monthly anomaly strength trajectories for the three districts contributing the largest malaria burden during anomalous periods, compared with the corresponding regional average anomaly strength. Anomaly strength ranges from 0 to 4 and represents the number of anomaly detection methods identifying a given district-month observation as anomalous. The figure illustrates the temporal relationship between district-level anomaly activity and broader regional behaviour.
}\label{fig:district_timeseries}
\end{figure*}

Temporal trajectories provide additional insight into the relationship between district-level and regional anomaly behaviour (Fig.~\ref{fig:district_timeseries}). Tamale exhibited considerable variability throughout the study period, with anomaly strengths fluctuating between normal conditions and the maximum consensus level. Strong anomalies were observed repeatedly during 2014--2017 and again during 2022--2023. Despite these fluctuations, the Tamale series closely tracked the Northern regional average, indicating that district-level behaviour exerted a substantial influence on the overall regional anomaly signal. The similarity between the two curves suggests that the dominant regional anomalies identified earlier were largely anchored in transmission dynamics occurring within Tamale and its surrounding urban catchment.

A different pattern emerged within the Ashanti Region. The anomaly trajectories for Kumasi and Atwima Nwabiagya were nearly indistinguishable from the regional average throughout the study period. Extended periods of anomaly strength equal to two were interrupted by repeated episodes reaching the strongest consensus category. Elevated anomaly activity became particularly evident after 2020, when several prolonged intervals reached anomaly strengths of three or four. The close correspondence between district and regional trajectories indicates that anomaly behaviour was spatially coherent across the Ashanti metropolitan corridor. Rather than being generated by a single district, the regional anomaly signal appears to reflect concurrent departures occurring across multiple neighbouring districts.

\begin{figure*}
\begin{minipage}[H]{\linewidth}
\centering
\includegraphics[width=\textwidth]{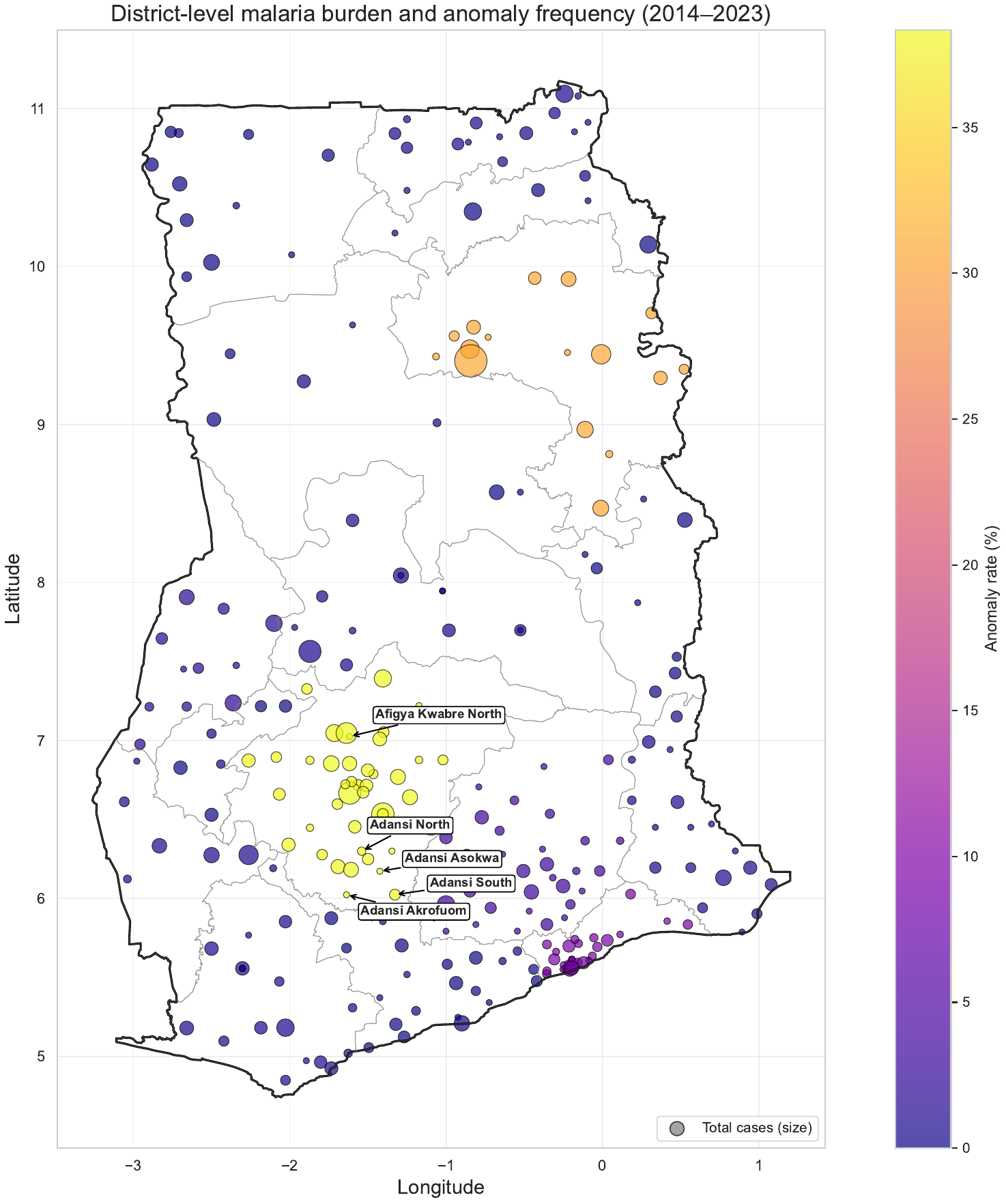} 
\end{minipage}
\caption{Bivariate representation of district-level malaria burden and anomaly frequency across Ghana during 2014--2023. Bubble size is proportional to cumulative malaria admissions, while colour indicates the percentage of months classified as anomalous ($S \geq 3$). Selected districts with particularly high anomaly rates are labelled. The figure distinguishes districts characterised by persistently high burden from those exhibiting frequent departures from expected transmission behaviour.
}\label{fig:bivariate_bubble_map}
\end{figure*}

The national distribution of anomaly activity is illustrated in Fig.~\ref{fig:bivariate_bubble_map}. A clear distinction is evident between districts characterised by high malaria burden and those characterised by high anomaly frequency. The largest bubbles, representing districts with substantial cumulative malaria burden, are concentrated within major population centres and regional capitals. In contrast, the highest anomaly rates, exceeding $35\%$, occur within a cluster of districts located primarily in the Ashanti belt. Several of these districts are highlighted around the Adansi and Afigya Kwabre areas, where anomaly rates are among the highest observed nationally despite more moderate cumulative case burdens. This pattern indicates that districts contributing strongly to anomaly detection are not necessarily those with the largest malaria burden. Instead, they are locations where malaria incidence deviates most frequently from expected local behaviour.

Spatial clustering of elevated anomaly rates is particularly evident in south-central Ghana. Districts exhibiting anomaly rates approaching $40\%$ form a geographically coherent cluster rather than isolated observations. Elsewhere, many districts with substantial malaria burden display anomaly rates below $10\%$, indicating relatively stable transmission dynamics despite high disease incidence. The spatial separation between burden and anomaly frequency demonstrates that the anomaly framework captures aspects of transmission variability that are not evident from case counts alone.

\begin{figure*}
\begin{minipage}[H]{\textwidth}
\centering
\includegraphics[width=\textwidth]{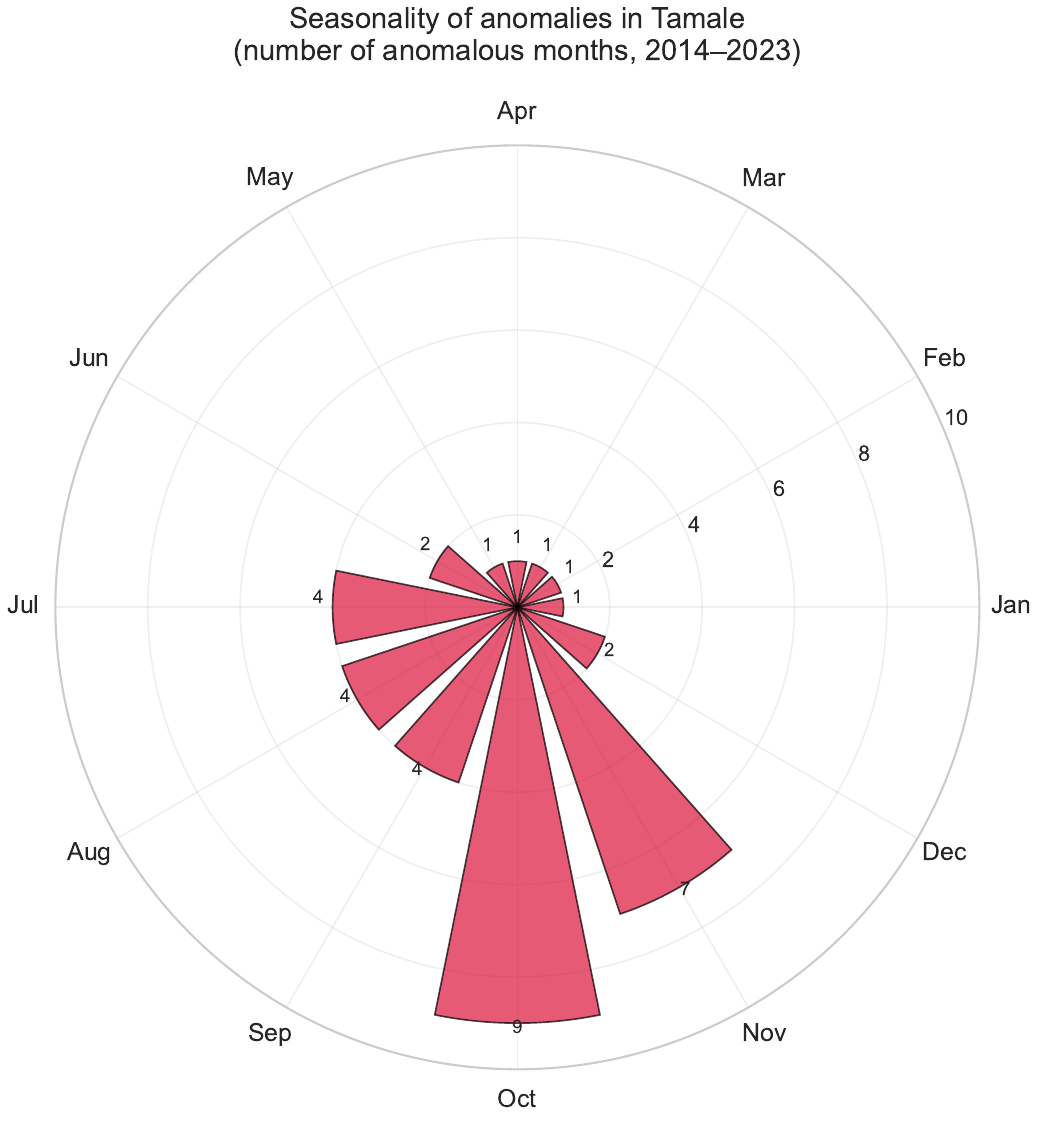} 
\end{minipage}
\caption{Seasonal distribution of anomalous months in Tamale District during 2014--2023. Radial bars indicate the number of months within each calendar month that were classified as anomalous by the consensus framework. The plot illustrates the seasonal concentration of anomaly occurrence within the district exhibiting the highest cumulative malaria burden during anomalous periods.
}\label{fig:seasonal_polar_plot}
\end{figure*}

Seasonal characteristics of the most burdened district are shown in Fig.~\ref{fig:seasonal_polar_plot}. Tamale experienced anomalous months throughout the calendar year, although anomaly occurrence was strongly concentrated during the latter half of the transmission cycle. October recorded the highest frequency, with nine anomalous months across the ten-year study period. November followed with seven occurrences, while July, August, and September each recorded four anomalous months. By contrast, January through June generally exhibited one or two anomalous months. The seasonal pattern aligns closely with the unimodal rainfall regime of northern Ghana, where peak malaria transmission commonly occurs during and immediately after the rainy season. The concentration of anomalies during September to November suggests that unusual transmission episodes were most likely to emerge near the seasonal peak rather than during low-transmission periods.

\begin{figure*}
\begin{minipage}[H]{\textwidth}
\centering
\includegraphics[width=\textwidth]{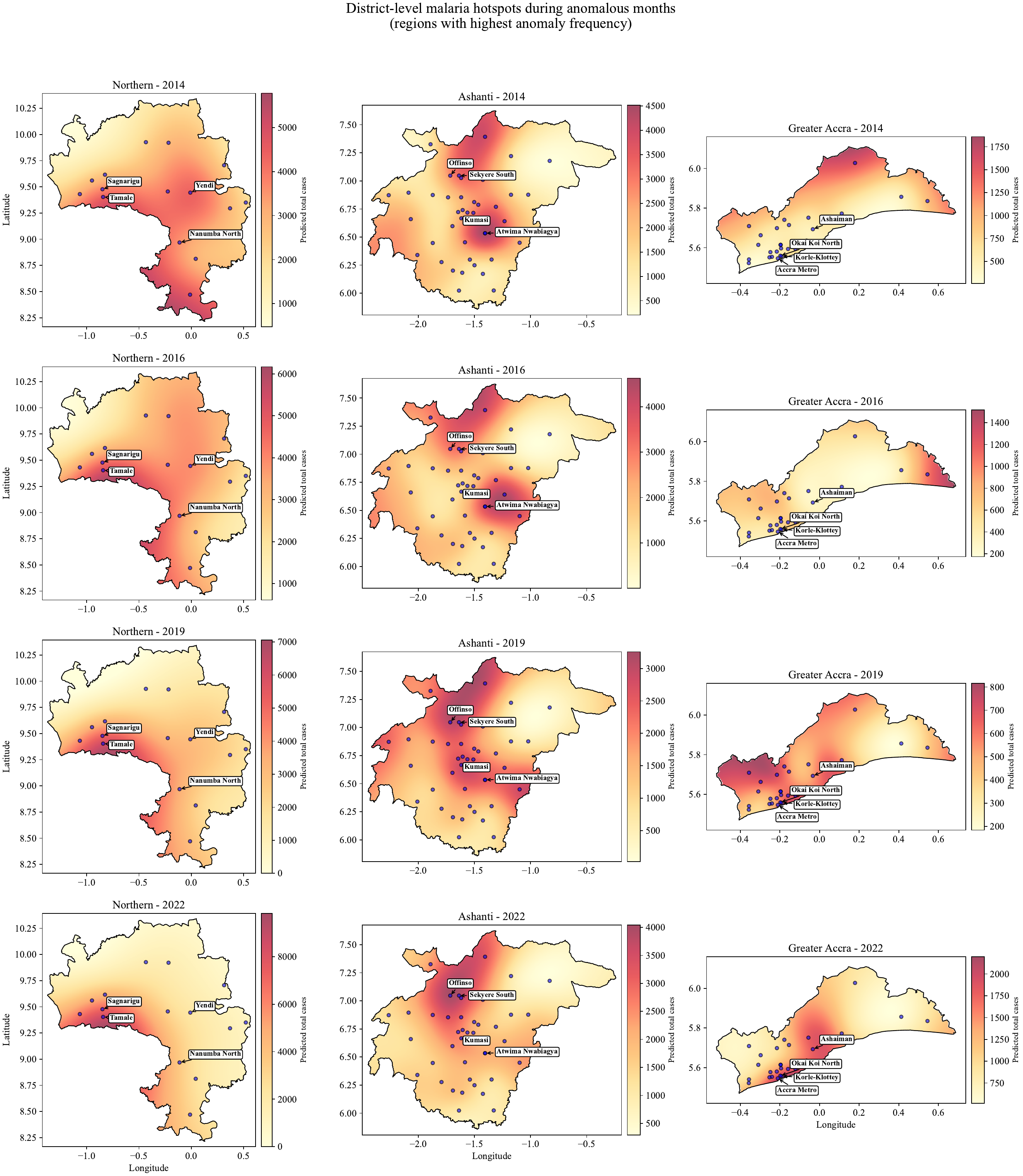} 
\end{minipage}
\caption{Spatial distribution of district-level malaria hotspots during anomalous months in Northern, Ashanti, and Greater Accra Regions for four representative years (2014, 2016, 2019, and 2022). Continuous surfaces were generated using radial basis function interpolation of the district malaria burden. Warmer colours indicate higher predicted malaria burden during anomalous periods. Labelled districts identify the principal hotspot locations within each region.
}\label{fig:district_hotspot_maps}
\end{figure*}

Figure~\ref{fig:district_hotspot_maps} provides a spatially explicit view of anomaly-associated malaria burden within the three regions exhibiting the highest anomaly frequencies. In Northern Region, hotspot intensity remained consistently centred on the Tamale metropolitan area across 2014, 2016, 2019, and 2022. Although the magnitude of predicted malaria burden varied between years, the geographical focus of the hotspot changed little. Secondary areas of elevated burden appeared intermittently toward Yendi and neighbouring districts, yet the dominant hotspot remained anchored within the central-western portion of the region.

Ashanti Region displayed a broader hotspot structure extending across Kumasi, Atwima Nwabiagya, Offinso, and Sekyere South. The location of maximum intensity shifted modestly between years, but the elevated burden remained concentrated within the same metropolitan and peri-metropolitan corridor. Spatial continuity between neighbouring districts was more pronounced than in the Northern Region, producing an extended hotspot zone rather than a single focal centre. The persistence of this corridor across all four representative years indicates that anomalous malaria activity repeatedly emerged within the same geographical setting.

Greater Accra exhibited the most spatially compact hotspot configuration. Elevated malaria burden remained concentrated within the Accra metropolitan area, particularly around Accra Metropolitan, Korle-Klottey, Okai Koi North, and Ashaiman. Although the centre of maximum intensity shifted modestly between years, the hotspot remained confined to a relatively small urban corridor. Compared with Northern and Ashanti Regions, the geographical extent of anomaly-associated burden was considerably smaller. Nevertheless, the repeated appearance of the hotspot across all representative years indicates persistent localised variability in malaria transmission within the metropolitan environment.

Comparison of Figs.~\ref{fig:district_bar_chart} and \ref{fig:bivariate_bubble_map} demonstrates that anomaly burden and anomaly frequency were spatially distinct. Tamale contributed the largest cumulative malaria burden during anomalous months, whereas the highest anomaly rates were concentrated within a cluster of districts in Ashanti Region, particularly around the Adansi and Afigya Kwabre areas. This contrast indicates that locations experiencing the greatest malaria burden were not necessarily those exhibiting the most persistent anomalous behaviour. Instead, burden and anomaly frequency highlighted different facets of malaria dynamics across the study area.

\subsection{Statistical Separation Between Anomalous and Normal Months}

To assess whether the consensus anomaly framework identified observations that were statistically distinct from routine transmission conditions, distributions of key epidemiological features were compared between anomalous months (anomaly strength $\geq 3$) and normal months (anomaly strength $\leq 2$). Differences were evaluated using the Mann--Whitney $U$ test, while Cohen's $d$ was used to quantify effect size (Table~\ref{tab:stat_validation}).

\begin{table*}[htbp]
\centering
\caption{Mann--Whitney $U$ test and Cohen's $d$ comparing anomalous (strength $\geq 3$) versus normal (strength $\leq 2$) months.}
\label{tab:stat_validation}
\begin{tabular}{lccc}
\toprule
\textbf{Feature} & \textbf{Mann--Whitney $U$} & \textbf{$p$-value} & \textbf{Cohen's $d$} \\
\midrule
Total cases   & 183142.5 & $5.31 \times 10^{-59}$ & 3.252 \\
Residual      & 140007.0 & $8.82 \times 10^{-17}$ & 1.383 \\
Region z-score & 145601.0 & $9.52 \times 10^{-21}$ & 1.245 \\
\bottomrule
\end{tabular}
\end{table*}

Results indicate highly significant differences between anomalous and normal months across all evaluated features. Total malaria cases exhibited the strongest separation, with a Mann--Whitney statistic of $U = 183142.5$ and an associated $p$-value of $5.31 \times 10^{-59}$. The corresponding Cohen's $d$ value of 3.252 represents an exceptionally large effect size, indicating that malaria case counts during anomalous months were substantially higher than those observed during normal months. The magnitude of this difference suggests that anomaly classification was strongly associated with periods of elevated malaria burden.

A similarly clear distinction was observed for the residual feature, which measures departures from the expected region-specific seasonal pattern. The Mann--Whitney test yielded $U = 140007.0$ with a $p$-value of $8.82 \times 10^{-17}$, while Cohen's $d$ reached 1.383. The combination of strong statistical significance and a large effect size indicates that anomalous months were characterised by markedly greater positive deviations from expected seasonal behaviour. This result demonstrates that the identified anomalies were not solely associated with high malaria burden, but also reflected substantial departures from the seasonal baseline within each region.

Region-specific standardised anomalies showed a comparable pattern. The region z-score differed significantly between the two groups ($U = 145601.0$, $p = 9.52 \times 10^{-21}$), with a Cohen's $d$ of 1.245. This large effect size indicates that anomalous months occurred substantially further from the regional mean than normal observations when expressed in standard deviation units. The result provides additional evidence that the anomaly framework preferentially identifies observations occupying the tails of the regional distribution rather than fluctuations occurring within the range of typical variability.

Taken together, the statistical tests demonstrate that anomalous and normal months represent clearly distinct populations within the surveillance record. The extremely small $p$-values indicate strong distributional separation across all three variables, while the large effect sizes confirm that these differences are substantial in magnitude. The strongest separation was observed for total malaria cases, although residuals and region-specific standardised deviations also exhibited pronounced differences. These findings provide quantitative support for the anomaly classifications identified by the consensus framework and confirm that anomalous months were associated with both elevated malaria burden and unusually large departures from expected regional transmission behaviour.

\section{Discussion}
\label{sec:disc}

The present study applied a consensus-based anomaly detection framework to ten years of routine malaria surveillance data from Ghana and identified a relatively small subset of region-month observations that departed substantially from expected transmission behaviour. Unlike conventional surveillance summaries, which primarily describe average burden and seasonal trends, the proposed framework was designed to identify observations that were statistically unusual relative to their own regional epidemiological context. Importantly, the detected anomalies should not be interpreted automatically as malaria outbreaks. Rather, they represent observations whose multivariate epidemiological characteristics differ substantially from historical expectations and therefore warrant closer epidemiological assessment. Such departures may arise from genuine increases in malaria transmission, unexpected persistence of transmission outside the usual seasonal period, changes in age-specific disease burden, alterations in healthcare utilisation, reporting inconsistencies, intervention effects, or short-term environmental perturbations. Consequently, the principal contribution of the framework is not the identification of regions with historically high malaria burden, but rather the prioritisation of unusual transmission patterns that may otherwise remain obscured within routine surveillance summaries.

One notable finding is that anomalous malaria activity was concentrated within a limited number of regions, particularly Ashanti and Northern. This concentration is unlikely to be explained solely by differences in malaria burden because the anomaly framework incorporated region-specific standardisation and seasonal residuals in addition to absolute case counts. Instead, the results suggest that regions characterised by large populations, heterogeneous ecological conditions, and pronounced seasonal transmission cycles may also experience greater temporal variability, increasing the likelihood of departures from expected behaviour. Previous studies have shown that malaria transmission in Ghana is strongly influenced by climatic variability, rainfall timing, temperature suitability, and local environmental conditions, all of which differ substantially between the forest, coastal and savannah ecological zones \cite{OhenebaDornyo2022,Asare2017}. The recurrent appearance of anomalies within Ashanti and Northern Regions may therefore reflect the interaction between high endemic transmission and episodic environmental or epidemiological disturbances rather than simply elevated malaria incidence. Importantly, these findings suggest that anomaly detection provides information that is complementary to disease burden, identifying instability in transmission dynamics rather than merely highlighting areas with persistently high case counts.

A particularly important observation emerging from the district-level analyses is that anomaly burden and anomaly persistence represent distinct epidemiological characteristics. Tamale contributed the largest cumulative malaria burden during anomalous months, whereas the highest anomaly frequencies were concentrated within a cluster of districts in Ashanti Region. This distinction suggests that the processes governing the magnitude of malaria transmission may differ from those responsible for repeated departures from expected behaviour. Districts with consistently high malaria burden may simply reflect large populations, sustained endemic transmission, or greater healthcare utilisation. Conversely, districts exhibiting recurrent anomalies may experience greater temporal instability resulting from fluctuating environmental conditions, intervention effectiveness, reporting practices, or local transmission dynamics. The ability to distinguish between consistently high transmission and repeated deviations from expected behaviour illustrates one of the principal advantages of anomaly detection over conventional burden-based surveillance metrics.

Seasonal structure provides additional insight into the detected anomalies. The concentration of anomalous activity during particular months indicates that the framework was sensitive to deviations occurring within established transmission seasons rather than merely identifying seasonal peaks themselves. This distinction is particularly important because malaria transmission in Ghana is inherently seasonal and follows different rainfall regimes across ecological zones. The northern savannah experiences a predominantly unimodal rainy season, whereas the forest and coastal zones exhibit bimodal rainfall patterns \cite{Asare2025,owusu2013changing}. Under these conditions, elevated malaria incidence during the rainy season is expected and should not automatically be regarded as anomalous. Instead, the proposed framework identifies months in which observed transmission differs substantially from the seasonal behaviour historically observed for that specific region. Such anomalies may therefore indicate transient environmental conditions, local operational challenges, unusual vector dynamics, changes in surveillance performance, or other factors that temporarily disrupt expected transmission patterns. Consequently, anomaly detection complements seasonal surveillance by distinguishing expected seasonal variation from statistically unusual seasonal behaviour.

The UMAP projection and inter-method agreement analyses provide important methodological insights into the characteristics of the detected anomalies. Strong anomalies occupied a well-defined region of the reduced feature space, indicating that these observations differed simultaneously across multiple epidemiological dimensions rather than along a single variable. This separation supports the rationale for adopting a multivariate anomaly detection framework rather than relying exclusively on univariate surveillance thresholds. Malaria anomalies frequently involve simultaneous changes in disease burden, seasonality, temporal persistence and demographic composition, none of which alone adequately characterises unusual transmission behaviour. Evaluating these complementary dimensions jointly therefore provides a more comprehensive representation of abnormal epidemiological patterns.

Differences in agreement among the four anomaly detection algorithms further illustrate the complexity of identifying unusual malaria transmission. The strong agreement between Isolation Forest and Elliptic Envelope indicates that many anomalies were sufficiently distinct to be recognised by both tree-based isolation and robust multivariate distance approaches. By contrast, Local Outlier Factor exhibited weaker agreement with the remaining algorithms because density-based methods primarily identify observations that are unusual within their immediate neighbourhood rather than globally unusual observations. This outcome suggests that multiple forms of epidemiological anomalies coexist within routine surveillance data. Some events represent globally extreme departures from historical behaviour, whereas others are only locally unusual within specific regions of the multivariate feature space. The consensus strategy reduces the influence of these algorithm-specific sensitivities by requiring agreement among multiple independent detectors before assigning high anomaly confidence, thereby improving robustness and reducing the likelihood of method-specific false positive detections.

An important implication of the present findings concerns the epidemiological interpretation of statistical anomalies. The proposed framework is designed primarily to identify abrupt or unexpected departures from historical transmission behaviour rather than gradual secular changes occurring over many years. Long-term shifts in malaria epidemiology, such as progressive increases or sustained reductions in regional transmission, are more appropriately investigated using complementary approaches including trend analysis, change-point detection, or probabilistic forecasting models. Anomaly detection therefore addresses a different surveillance objective by identifying observations that deviate unexpectedly from historical expectations irrespective of whether the underlying long-term trend is increasing or decreasing. This distinction is important because different analytical approaches answer different epidemiological questions and should be regarded as complementary rather than competing methodologies.

From an operational perspective, the findings demonstrate the potential value of anomaly detection as an additional analytical layer within routine malaria surveillance rather than as a replacement for existing surveillance systems. Current surveillance programmes primarily rely on temporal trends, predefined thresholds and routine descriptive reporting. While these approaches remain fundamental components of malaria control programmes, they may overlook statistically unusual observations that do not exceed conventional alert thresholds but nevertheless warrant epidemiological review. By prioritising observations that depart substantially from expected regional behaviour, consensus anomaly detection can support surveillance officers in identifying records that merit further investigation. Depending on the epidemiological context, such investigations may include verification of surveillance data quality, assessment of intervention coverage, review of diagnostic and treatment commodity availability, intensified vector surveillance, or targeted field investigations. The framework therefore functions as a decision-support tool that strengthens epidemiological situational awareness while recognising that subsequent public health responses remain dependent upon local operational capacity, surveillance infrastructure and programme priorities.

Several limitations should be acknowledged when interpreting the present findings. First, the analysis relied on routinely collected surveillance data aggregated at regional and district levels. Although DHIMS2 provides extensive spatial and temporal coverage, reporting completeness, diagnostic practices, and healthcare utilisation may vary across locations and years, potentially influencing the observed anomaly patterns. Second, anomaly detection identifies statistical departures from expected behaviour but cannot determine the underlying causal mechanisms responsible for those departures. Consequently, the detected anomalies should be interpreted as surveillance signals requiring further epidemiological investigation rather than definitive evidence of malaria outbreaks or intervention failures. Third, the selected feature set was intentionally restricted to variables derived from routine surveillance records. Incorporating environmental covariates such as rainfall, temperature, vegetation indices, flooding indicators, vector abundance, intervention coverage, and socioeconomic factors may improve interpretation of anomaly events and facilitate investigation of their underlying drivers. Fourth, although the proposed framework captures both spatial and temporal variation by analysing observations collected across multiple regions over time, anomaly detection is performed independently for each region and therefore does not explicitly model spatial dependence among neighbouring regions. Malaria transmission is known to exhibit spatial autocorrelation arising from shared environmental conditions, population mobility, vector ecology, and healthcare accessibility. Consequently, the present framework identifies observations that deviate from each region's own historical transmission characteristics rather than departures relative to neighbouring regions. Explicit incorporation of spatial dependence may therefore improve the identification of geographically coordinated transmission anomalies and provide additional insight into the spatial processes underlying malaria dynamics. Fifth, the present analysis was designed as a retrospective assessment of historical surveillance data to identify statistically unusual transmission patterns over the study period. Although this retrospective framework is appropriate for evaluating historical epidemiological behaviour, operational deployment within routine malaria surveillance would require sequential updating of summary statistics and model calibration using only information available up to each reporting period. Prospective implementation under such real-time conditions represents an important next step towards integrating anomaly detection into routine surveillance workflows. Finally, the present study did not evaluate whether earlier identification of statistical anomalies translates directly into improved public health outcomes. Assessing the operational effectiveness of anomaly-informed surveillance will require prospective implementation and evaluation within routine malaria control programmes.

Future work should extend the proposed framework in several complementary directions. Integrating climate reanalysis products, satellite-derived environmental indicators, entomological surveillance data, intervention records, and human mobility information within a unified spatiotemporal framework would facilitate investigation of the environmental and operational drivers underlying detected anomalies while improving epidemiological interpretation. An important methodological extension is the incorporation of explicit spatial dependence within the anomaly detection framework itself. Rather than evaluating each region independently, future models could integrate neighbourhood information through spatial lag variables, adjacency-based feature engineering, or graph representations of regional connectivity, enabling anomalies to be identified with respect to both local historical behaviour and contemporaneous conditions in neighbouring regions. Such extensions would facilitate the identification of geographically coordinated transmission events and improve understanding of the spatial propagation of malaria anomalies. Another promising direction is the incorporation of simulation-based evaluation using synthetic or model-generated malaria time series. Carefully designed simulation studies could provide controlled environments for examining the sensitivity and robustness of consensus anomaly detection algorithms under different transmission regimes, seasonal structures, and anomaly scenarios while complementing validation based on routine surveillance observations. In addition, prospective implementations based on rolling or sequential model updating would enable assessment of algorithm performance under operational surveillance conditions where future observations are unavailable at the time of analysis. Finally, combining anomaly detection with probabilistic forecasting represents another important avenue for methodological development. Forecasting models estimate expected malaria transmission trajectories together with associated uncertainty, whereas anomaly detection identifies observations that depart systematically from those expectations. Integrating these complementary methodologies within a unified spatiotemporal framework has the potential to provide a more comprehensive surveillance system capable of simultaneously anticipating future malaria transmission, identifying statistically unusual epidemiological behaviour, and characterising the spatial evolution of transmission dynamics.

\section{Conclusion} \label{sec:conc}

This study presented a national-scale assessment of malaria transmission anomalies in Ghana using a consensus anomaly detection framework applied to routine surveillance records spanning 2014 to 2023. By combining multiple unsupervised learning algorithms with a set of epidemiologically informed features, the framework was designed to identify observations that deviated from expected regional transmission behaviour rather than simply highlighting periods of high malaria burden. In doing so, the study addressed an important gap in malaria surveillance by providing a systematic approach for detecting unusual transmission dynamics within large-scale routine health datasets.

The findings demonstrate that malaria anomalies in Ghana exhibit clear spatial, temporal, and seasonal organisation. Rather than occurring randomly across the surveillance record, anomalous transmission was concentrated within specific regions, districts, and periods of the year, indicating the presence of geographically structured departures from expected epidemiological behaviour. The analysis further revealed that the strongest anomaly signals were associated with identifiable regional and district-level hotspots that persisted through time, suggesting that unusual transmission dynamics are repeatedly expressed within particular transmission environments. These observations highlight the importance of examining malaria variability beyond conventional measures of burden alone.

An additional contribution of the study is the demonstration that anomaly burden and anomaly frequency represent distinct dimensions of malaria dynamics. Districts contributing the largest malaria burden during anomalous periods were not always those exhibiting the most persistent anomalous behaviour. This distinction would have remained largely obscured under traditional surveillance approaches focused solely on case counts. The results therefore illustrate how anomaly detection can provide complementary information regarding transmission stability, variability, and departure from expected local conditions.

The study also provides evidence that anomalous months constitute a statistically distinct component of the surveillance record. Strong anomalies exhibited coherent multivariate characteristics, occupied separate regions of the epidemiological feature space, and remained identifiable across multiple independent detection algorithms. The convergence of spatial, temporal, statistical, and methodological evidence supports the interpretation that the detected anomalies represent genuine departures from routine transmission behaviour rather than artefacts arising from individual analytical techniques.

Beyond the specific findings for Ghana, this work demonstrates the value of consensus anomaly detection as a practical surveillance tool for monitoring endemic diseases. The framework leverages routinely collected health data, requires no prior outbreak labels, and can be implemented within existing surveillance infrastructures. As national health information systems continue to expand in coverage and quality, anomaly-based approaches offer an opportunity to complement conventional monitoring strategies by identifying unusual epidemiological events that may warrant further investigation.

Taken together, the study shows that routine malaria surveillance data contain substantially more information than is captured through standard burden assessments alone. By shifting attention from absolute disease counts to departures from expected transmission behaviour, the proposed framework provides a new perspective for understanding malaria dynamics and establishes a foundation for future surveillance systems that integrate anomaly detection, forecasting, environmental monitoring, and early-warning capabilities within a unified analytical framework.

\section*{CRediT authorship contribution statement}

\textbf{T. Ansah-Narh:} Conceptualisation, Formal analysis, Validation, Methodology, Writing – original draft, and Writing – review and editing. \\

\textbf{Y. Asare Afrane:} Data curation, Funding acquisition, Project administration, Supervision, Resources, and Writing – review and editing. \\


\section*{Declaration of competing interest}
The authors confirm that there are no financial, professional, or personal relationships that could be construed as potential conflicts of interest in relation to this work.

\section*{Data availability} 

The data used in this study consist of monthly district-level malaria admission records for two population groups: children under five years of age and individuals aged five years and above. The dataset was obtained from the Disease Surveillance Department of the Ghana Health Service. Owing to data-sharing restrictions, confidentiality agreements, and institutional data protection policies, the raw data are not publicly accessible. Researchers interested in accessing the data may submit a formal request to the Ghana Health Service, which will evaluate applications on a case-by-case basis, subject to ethical approval and administrative authorisation. All analyses were conducted using anonymised and aggregated data, and no personally identifiable information was accessed or utilised during the study.

\section*{Funding}
Funding for this study was provided by the Bill \& Melinda Gates Foundation (INV-047051) through the West Africa Mathematical Modelling Capacity Development (WAMCAD) initiative, with additional support from the National Institutes of Health under Grant D43 TW011513.

\section*{Acknowledgements}

The authors sincerely acknowledge the support of the West Africa Mathematical Modelling Capacity Development (WAMCAD) programme for its institutional, technical, and logistical assistance throughout this research. WAMCAD has been instrumental in strengthening regional expertise in mathematical and computational modelling, particularly in the study of malaria and other neglected tropical diseases.
The authors are also grateful to the Ghana Health Service for providing access to the malaria surveillance data used in this study and for the valuable technical guidance that helped shape the analytical approaTheadopted.
Finally, the authors express their appreciation to the anonymous reviewers for their constructive comments and insightful recommendations. Their feedback significantly enhanced the clarity, methodological robustness, and overall quality of the manuscript.

\appendix
\section{Principal Component Analysis of Engineered Features: Mathematical Formulation and Interpretation}
\label{app:pca}

This appendix provides the mathematical details of the Principal Component Analysis (PCA) used to investigate the correlation structure among the nine engineered features described in Section~\ref{subsec:feature_engineering}. PCA was employed as an exploratory technique to identify dominant modes of variability, quantify feature redundancy, and facilitate interpretation of the correlation circle shown in Fig.~\ref{fig:correlation_circle}. The derivation presented here establishes the relationship between the standardised data matrix, covariance matrix, eigenvalue decomposition, and principal component loadings.

\subsection{Notation and Data Standardisation}

Let

\begin{equation}
\mathbf{X}
=
\left[x_{ij}\right]
\in \mathbb{R}^{n\times p},
\label{eq:pca_data_matrix}
\end{equation}

denote the feature matrix, where \(n=1908\) represents the number of region--month observations and \(p=9\) denotes the number of engineered features.

Because the variables are measured on different scales, PCA was performed on a standardised feature matrix. Each feature was centred and scaled according to

\begin{equation}
z_{ij}
=
\frac{x_{ij}-\bar{x}_j}{s_j},
\qquad
i=1,\ldots,n,\;
j=1,\ldots,p,
\label{eq:pca_standardisation}
\end{equation}

where

\begin{equation}
\bar{x}_j
=
\frac{1}{n}
\sum_{i=1}^{n}
x_{ij},
\label{eq:pca_mean}
\end{equation}

is the sample mean of feature \(j\), and

\begin{equation}
s_j
=
\sqrt{
\frac{1}{n-1}
\sum_{i=1}^{n}
\left(x_{ij}-\bar{x}_j\right)^2
},
\label{eq:pca_sd}
\end{equation}

is the corresponding sample standard deviation.

Equation~\eqref{eq:pca_standardisation} transforms each feature to zero mean and unit variance, ensuring that variables with large numerical magnitudes do not dominate the PCA solution. The resulting standardised matrix is denoted by

\begin{equation}
\mathbf{Z}
=
\left[z_{ij}\right]
\in
\mathbb{R}^{n\times p}.
\label{eq:pca_zmatrix}
\end{equation}

\subsection{Covariance Matrix and Eigenvalue Decomposition}

The sample covariance matrix of the standardised data is computed as

\begin{equation}
\mathbf{S}
=
\frac{1}{n-1}
\mathbf{Z}^{\top}\mathbf{Z},
\label{eq:pca_covariance}
\end{equation}

\noindent  where
\(
\mathbf{S}
\in
\mathbb{R}^{p\times p}
\)
is symmetric and positive semi-definite.

PCA seeks orthogonal directions that maximise the variance of the projected data. These directions are obtained from the eigenvalue problem

\begin{equation}
\mathbf{S}\mathbf{v}_k
=
\lambda_k \mathbf{v}_k,
\qquad
k=1,\ldots,p,
\label{eq:pca_eigenproblem}
\end{equation}

\noindent  where \(\lambda_k\) denotes the \(k\)-th eigenvalue and \(\mathbf{v}_k\) is the associated eigenvector.

The eigenvalues are ordered as

\begin{equation}
\lambda_1
\ge
\lambda_2
\ge
\cdots
\ge
\lambda_p
\ge
0,
\label{eq:pca_eigen_order}
\end{equation}

\noindent such that the first principal component corresponds to the direction of maximum variance.

The \(k\)-th principal component score vector is obtained by projecting the standardised observations onto the eigenvector \(\mathbf{v}_k\),

\begin{equation}
\mathbf{t}_k
=
\mathbf{Z}\mathbf{v}_k,
\label{eq:pca_scores}
\end{equation}

\noindent  where
\(
\mathbf{t}_k \in \mathbb{R}^{n}
\)
contains the principal component scores for all observations.
Equation~\eqref{eq:pca_scores} provides the coordinates of each observation in the reduced-dimensional PCA space.

\subsection{Variance Explained}

The total variance contained in the standardised dataset equals the trace of the covariance matrix,

\begin{equation}
\mathrm{tr}(\mathbf{S})
=
\sum_{k=1}^{p}\lambda_k.
\label{eq:pca_total_variance}
\end{equation}

\noindent 
The proportion of variance explained (PVE) by the \(k\)-th principal component is therefore

\begin{equation}
\mathrm{PVE}_k
=
\frac{\lambda_k}
{\sum_{j=1}^{p}\lambda_j}.
\label{eq:pca_pve}
\end{equation}

\noindent 
Equation~\eqref{eq:pca_pve} was used to compute the reported contributions of the first two principal components, namely 46.2\% and 15.4\%, respectively.

The cumulative proportion of explained variance for the first \(m\) components is

\begin{equation}
\mathrm{CPVE}_m
=
\sum_{k=1}^{m}
\mathrm{PVE}_k.
\label{eq:pca_cpve}
\end{equation}

\noindent 
Using Eq.~\eqref{eq:pca_cpve}, the first two components explain 61.6\% of the total variance.

\subsection{Loadings and the Correlation Circle}

The eigenvectors obtained from Eq.~\eqref{eq:pca_eigenproblem} define the principal component loading matrix,

\begin{equation}
\mathbf{V}
=
\left[
\mathbf{v}_1,
\mathbf{v}_2,
\ldots,
\mathbf{v}_p
\right].
\label{eq:pca_loading_matrix}
\end{equation}

\noindent  Because the eigenvectors are orthonormal,

\begin{equation}
\mathbf{V}^{\top}\mathbf{V}
=
\mathbf{I},
\label{eq:pca_orthogonality}
\end{equation}

\noindent where \(\mathbf{I}\) denotes the identity matrix.

The coordinates of feature \(j\) in the correlation circle are computed as

\begin{equation}
\ell_{jk}
=
\sqrt{\lambda_k}\,v_{jk},
\label{eq:pca_loading_coordinate}
\end{equation}

\noindent 
where \(v_{jk}\) denotes the \(j\)-th element of eigenvector \(\mathbf{v}_k\).

Consequently, the coordinates of feature \(j\) in the PC1--PC2 plane are

\begin{equation}
\left(
\ell_{j1},
\ell_{j2}
\right)
=
\left(
\sqrt{\lambda_1}v_{j1},
\sqrt{\lambda_2}v_{j2}
\right).
\label{eq:pca_correlation_circle}
\end{equation}

\noindent 
Equation~\eqref{eq:pca_correlation_circle} forms the basis of the correlation circle shown in Fig.~\ref{fig:correlation_circle}. Features with similar directions exhibit positive correlation, orthogonal features are approximately uncorrelated, and features pointing in opposite directions exhibit negative correlation.

\subsection{Singular Value Decomposition and Computational Complexity}

In practice, PCA was computed using the singular value decomposition (SVD) of the standardised matrix \(\mathbf{Z}\),

\begin{equation}
\mathbf{Z}
=
\mathbf{U}
\mathbf{\Sigma}
\mathbf{V}^{\top},
\label{eq:pca_svd}
\end{equation}

\noindent 
where

\[
\mathbf{U}\in\mathbb{R}^{n\times p},
\qquad
\mathbf{\Sigma}\in\mathbb{R}^{p\times p},
\qquad
\mathbf{V}\in\mathbb{R}^{p\times p}.
\]

The diagonal matrix

\[
\mathbf{\Sigma}
=
\mathrm{diag}
(\sigma_1,\sigma_2,\ldots,\sigma_p)
\]

\noindent contains the singular values ordered as

\[
\sigma_1
\ge
\sigma_2
\ge
\cdots
\ge
\sigma_p
\ge
0.
\]

\noindent  Substituting Eq.~\eqref{eq:pca_svd} into Eq.~\eqref{eq:pca_covariance} yields

\begin{equation}
\mathbf{S}
=
\frac{1}{n-1}
\mathbf{V}
\mathbf{\Sigma}^{2}
\mathbf{V}^{\top},
\label{eq:pca_svd_covariance}
\end{equation}

\noindent  from which the eigenvalues are obtained as

\begin{equation}
\lambda_k
=
\frac{\sigma_k^2}{n-1}.
\label{eq:pca_singular_eigen}
\end{equation}

\noindent  Equation~\eqref{eq:pca_singular_eigen} establishes the equivalence between the SVD and eigenvalue formulations of PCA.

The dominant computational cost arises from the SVD in Eq.~\eqref{eq:pca_svd}, requiring

\begin{equation}
O(np^2)
\label{eq:pca_complexity}
\end{equation}

\noindent  floating-point operations when \(n \gg p\). For the present study (\(n=1908\), \(p=9\)), the computational burden is negligible and memory requirements scale as \(O(np)\).

\subsection{Interpretation of Figure~\ref{fig:correlation_circle}}

The correlation circle in Figure~\ref{fig:correlation_circle} was generated using the coordinates defined by Eq.~\eqref{eq:pca_correlation_circle}. The first two eigenvalues obtained from Eq.~\eqref{eq:pca_eigenproblem} were

\[
\lambda_1 = 4.16,
\qquad
\lambda_2 = 1.38,
\]

\noindent  which correspond to proportions of explained variance computed using Eq.~\eqref{eq:pca_pve}. Together, these two components account for 61.6\% of the total variance according to Eq.~\eqref{eq:pca_cpve}.

The tight clustering of \texttt{total\_cases}, \texttt{under5}, and \texttt{over5} reflects similar loading coordinates in Eq.~\eqref{eq:pca_loading_coordinate}, confirming strong positive correlation among these variables. Conversely, the near-orthogonal positioning of \texttt{month\_sin} and \texttt{month\_cos} relative to the case-related variables indicates that the seasonal encodings represent an independent temporal signal. This interpretation, derived directly from Eqs.~\eqref{eq:pca_loading_coordinate} and \eqref{eq:pca_correlation_circle}, supports the retention of all nine engineered features for anomaly detection.


 \bibliographystyle{elsarticle-num-names.bst} 
 \bibliography{refs}

\end{document}